\documentclass[10pt,journal,compsoc]{IEEEtran}
\ifCLASSOPTIONcompsoc
  \usepackage[nocompress]{cite}
\else
  \usepackage{cite}
\fi
\ifCLASSINFOpdf
\else
\fi
\usepackage{algpseudocode}
\usepackage{algorithm}
\usepackage{graphicx}
\usepackage{balance}

\usepackage{longtable}
\usepackage{tabularx,booktabs}
\usepackage{subfiles}
\usepackage{array}
\begin{document}
%


\title{Deep Bayesian Image Set Classification: A Defence Approach against Adversarial Attacks}
%
%
%
%

\author{Nima~Mirnateghi,
        Syed Afaq Ali~Shah,
        and~Mohammed~Bennamoun
\IEEEcompsocitemizethanks{\IEEEcompsocthanksitem N. Mirnateghi and S. Shah are with the Discipline of Information Technology, Murdoch University, Perth, Australia.\protect\\
\IEEEcompsocthanksitem M. Bennamoun is with the Department of Computer Science and Software Engineering, University of Western Australia, Crawley, WA, Australia.}
\thanks{Manuscript received December 22, 2020; revised.}}

%
%

\markboth{Journal of \LaTeX\ Class Files,~Vol.~14, No.~8, August~2015}%
{Shell \MakeLowercase{\textit{et al.}}: Bare Demo of IEEEtran.cls for Computer Society Journals}
%



\IEEEtitleabstractindextext{%
\begin{abstract}
Deep learning has become an integral part of various computer vision systems in recent years due to its outstanding achievements for object recognition, facial recognition, and scene understanding. However, deep neural networks (DNNs) are susceptible to be fooled with nearly high confidence by an adversary. In practice, the vulnerability of deep learning systems against carefully perturbed images, known as adversarial examples, poses a dire security threat in the physical world applications. To address this phenomenon, we present, what to our knowledge, is the first ever image set based adversarial defence approach. Image set classification has shown an exceptional performance for object and face recognition, owing to its intrinsic property of handling appearance variability. We propose a robust deep Bayesian image set classification as a defence framework against a broad range of adversarial attacks. We extensively experiment the performance of the proposed technique with several voting strategies. We further analyse the effects of image size, perturbation magnitude, along with the ratio of perturbed images in each image set. We also evaluate our technique with the recent state-of-the-art defence methods, and single-shot recognition task. The empirical results demonstrate superior performance on CIFAR-10, MNIST, ETH-80, and Tiny ImageNet datasets.
\end{abstract}

}

\maketitle

\IEEEdisplaynontitleabstractindextext

%
\IEEEpeerreviewmaketitle

\IEEEraisesectionheading{\section{Introduction}\label{sec:introduction}}

%
%
%
%
\IEEEPARstart{D}{eep} 
 Learning systems \cite{Lecun2015} have seen exceptional advances in the domain of computer vision, and have achieved a significant breakthrough for several tasks, such as object recognition, object detection, and segmentation to name a few \cite{krizhevsky2012imagenet,chen_semantic,He_ResNet}. The continuous advents of Deep Convolutional Neural Networks (CNNs) in numerous Artificial Intelligence (AI) based systems, such as autonomous cars \cite{akhtar2018threat}, robotics \cite{giusti2015machine}, and medical sciences \cite{esteva2017dermatologist}, confirm their paramount importance in the present and the coming future \cite{akhtar2018threat}. Despite their promising performance in a number of real-world applications, carefully perturbed images, which are imperceptible to human eyes, can easily cause the networks to misclassify with high confidence \cite{ szegedy2013intriguing}. In order to mitigate such security threats to deep learning systems, various defence mechanisms have been introduced in the literature.

The vulnerability of deep learning models to misclassification by image perturbation, named as adversarial examples, was first acknowledged in the study by Szegedy et al. \cite{szegedy2013intriguing}. The finding was a turning point for researchers to explore new implementations of adversarial attacks by generating perturbations on a sample of images to fool various deep learning  architectures \cite{moosavi2016deepfool,madry2017towards,su2019onepixel,moosavi2017universal,goodfellow2014explaining, mopuri2017fast}. Moreover, the robustness of adversarial attacks on different viewpoints and camera angles for real world objects has also been demonstrated in some studies\cite{athalye2018synthesizing, roadsignEykholt}. Adversarial attacks can indeed jeopardise many of our daily tasks in the physical world.
Several studies have highlighted the effect of adversarial perturbations on the misclassification of road signs by autonomous driving vehicles \cite{roadsignEykholt}, robotic vision\cite{RoboticvisionMelis2017}, healthcare systems \cite{finlayson2018adversarial}, cyberspace \cite{papernot2017cyberspace}, as well as cell-phone cameras \cite{kurakin2016adversarial_phone}.

\begin{figure}
  \centering

    \includegraphics[width=0.48\textwidth]{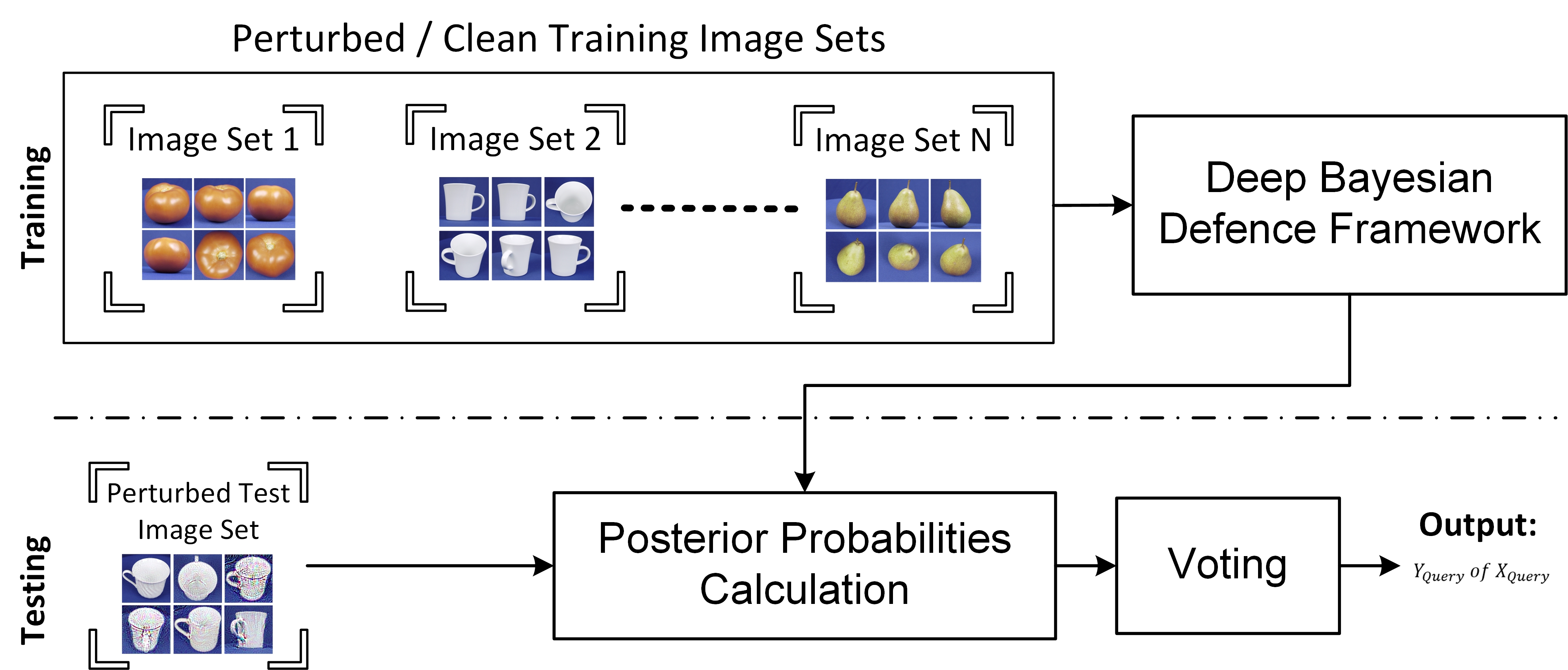}
  \caption{The Block Diagram of our Proposed Defence Approach against Adversarial Attacks.}
  \label{Figure_1}
\end{figure}

In response, this vulnerability of deep neural networks has attracted the attention of many researchers to posit defence mechanisms. An overview of the literature suggest that most of these theories focus on adversarial training \cite{goodfellow2014explaining,wang2019bilateral,shafahi2019adversarial,madry2017towards,moosavi2016deepfool,wong2020fast,tramer2017ensemble}, some on detection systems \cite{feinman2017detecting,li2017adversarial,grosse2017statistical}, and others on modifying the network layers \cite{Papernot_Dissiltation}. Even though the majority of the current defence systems have arguably demonstrated a degree of robustness against adversarial examples, they are still prone to be defeated \cite{carlini2017adversarial}.

This paper proposes a novel image set classification based approach to defend against adversarial attacks. In recent years, image set classification has emerged as an accurate classification technique for object recognition, facial recognition, and security surveillance \cite{zhao2019review,hayat2015,shah2016iterative,nadeem2018real}. Instead of classifying objects from single-shot images, as in traditional methods, image set classification recognises objects from a group of images that belong to one class. Image set classification can provide an opportunity for classifiers to learn about a wider range of appearance variations among images, such as different viewpoints and poses \cite{nadeem2018real}. The rich information provided by each image set brings many advantages to mitigate the weaknesses of traditional recognition methods in the presence of noise or distortions \cite{nadeem2018real}, allowing image set algorithms to outperform. These additional information can help deep learning models to gain extra knowledge, which may be impossible to achieve by the traditional image recognition tasks.

The promising performance of image set classification for imaging domain, its ability to handle appearance variations as well as its success in deep learning applications motivates the need for image set classification as a strong defence technique against a wide range of adversaries. The  contribution of  this  paper  can  be  summarised  as follows:
\begin{itemize}
    \item A novel image set classification based approach for defence against adversarial attacks is proposed. To the best of our knowledge, \textbf{this is the first ever image set based defence approach (Section \ref{method})}.
    \item The proposed method learns the patterns of image sets from a deep Bayesian neural network.
    \item We demonstrate the performance of the proposed image set based technique through extensive experiments using different classification voting strategies, including majority voting, exponential weighted voting as well as soft voting (Section \ref{experimental result}).
    \item The reported experimental results demonstrate that the proposed deep Bayesian image set classification network achieves a superior performance on benchmark datasets, and outperforms state-of-the-art defence techniques (Section \ref{StateOfTheArt}).
    \item In order to assess the generalisability of our propsed technique, the effect of perturbation magnitude, image size, and the ratio of perturbed images in each image set is evaluated (Section \ref{ablative}). We also assess the performance of our technique with single-shot recognition task. 
\end{itemize}
The rest of the paper is organised as follows. Related work is provided in the next section. The proposed methodology is described in Section \ref{method}. Experimental results are reported in Section \ref{experimental result} and the paper is concluded in Section \ref{conclusion}.

\section{Related Work} \label{related_work}
The common characteristics of adversarial threat models that are reported in the literature are the level of adversary's access to deep learning systems and the transferability of adversarial examples \cite{akhtar2018threat, YuanAdversarialExamples2019, zhao2019review, zhang2019adversarial}.
The adversary's knowledge is classified based on the amount of knowledge that the attackers have about their target. For illustration, in white-box settings, attackers construct adversarial examples in reference to their complete knowledge of the target model, including the network architecture, the number of layers, weights, the values of the hyperparameters, and sometimes the training data. On the other hand, a black-box attack assumes that the adversary does not have any access or knowledge of their target network, except the model output labels. It is also argued that many online machine learning services are vulnerable to black-box attacks \cite{YuanAdversarialExamples2019}. Such a property of adversarial examples motivated Papernot et al. \cite{papernot2017cyberspace} to discover the transferability of black-box attacks to other networks, which is also investigated in other studies \cite{liu2016delving,tramer2017space,goodfellow2014explaining}. From the defenders' perspective, the transferability of adversarial examples is of utmost importance. A robust defence strategy in the physical world should be able to demonstrate superior performance against attacks generated from other deep learning models in a black-box setting. In the rest of this section, an overview of the recent adversarial attacks is given, followed by the current theorised countermeasures that are available in the literature. We also study some deep learning applications for image set classification. 

\subsection{Adversarial Attacks}
\subsubsection{Fast Gradient Sign Method (FGSM)}
Since the discovery of adversarial examples as one of the intriguing vulnerabilities of neural networks \cite{szegedy2013intriguing}, many speculated that the non-linearity of deep learning networks could be the main cause of being fooled. 
However, Goodfellow et al. \cite{goodfellow2014explaining}, hypothesised that the linear behaviour of the models in high-dimensional data manifolds is the underlying reason behind adversarial perturbations. Based on this assumption, the authors developed FGSM attack to fool neural networks. The attack was generated by linearising the cost function to efficiently compute the gradient descent using back-propagation. Let $y$ be the target label, $I_{c}$ the input clean image, $\theta$ the network parameters, the adversarial example $I^{adv}$ is generated by calculating the gradient cost function $\nabla \mathcal J (.., .. , ..)$ using the $sign$ function with an $\epsilon$ norm value or perturbation magnitude.
 
\begin{equation} 
I^{adv} = {I}_{c} + \epsilon \cdot sign (\nabla \mathcal{J} (\theta, \mathbf {I}_{c}, y))
\end{equation}

\subsubsection{Deep Fool}
Moosavi-Dezfooli et al. \cite{moosavi2016deepfool} proposed an iterative method to generate perturbations based on the minimal norm. For each iteration, DeepFool computes a small perturbation vector. The algorithm then perturbs the image and computes the distance from the resulting image to the decision boundary of the classifier. This process is repeated until the label of the perturbed image is changed \cite{akhtar2018threat}. Moosavi-Dezfooli et al. \cite{moosavi2016deepfool} observed that their algorithm calculates a smaller perturbation size compared to FGSM \cite{goodfellow2014explaining}, leading to lower computational costs. Moreover, Moosavi-Dezfooli et al. \cite{moosavi2016deepfool} performed an experiment to assess the robustness of deep learning classifiers. It was shown that fine-tunning deep neural networks on adversarial examples generated by DeepFool substantially improved the robustness of the networks as well as the accuracy after training a few additional epochs. As a result, the authors concluded that combining training data with DeepFool adversarial examples is an effective method to build robust networks on large-scale datasets.

\subsubsection{Projected Gradient Descent (PGD)}
The robustness of adversarial examples generally requires a high value of perturbation size to effectively fool neural networks \cite{madry2017towards}. Projected Gradient Descent (PGD) attack \cite{madry2017towards} is known as a first-order adversary that is an iterative variant of FGSM\cite{goodfellow2014explaining}. The attack aims to generate robust adversarial examples with small perturbation size. Madry et al. \cite{madry2017towards} also highlighted that training a deep neural network with PGD adversarial examples enhances the network robustness. 

\begin{equation} \label{PGD}
    I^{t+1}=\Pi _{S_{I}}(I^{t}+\epsilon \cdot sign(\nabla _{I}L(\theta, I^{t},y )))
\end{equation}
Given the original image $I$, several PGD steps as shown in Equation \ref{PGD} are performed to obtain the projected perturbed image $I^{t+1}$.

\subsubsection{Universal Adversarial Perturbation (UAP) }
Moosavi-Dezfooli et al. \cite{moosavi2017universal} further extended their work on DeepFool \cite{moosavi2016deepfool} to develop a universal attack, which causes deep neural networks to misclassify with a high probability \cite{moosavi2017universal}. The adversarial examples in this method are generated using only a small sample of input images instead of the whole dataset. It was shown that universal adversarial perturbations generalise well both on fooling groups of images, as well as various deep learning architectures, including ResNet\cite{He_ResNet}, CaffeNet \cite{jia2014caffe}, VGG16 \cite{simonyan2014very} and VGG19 \cite{simonyan2014very}. 
These findings indicate the transferability and image-agnostic properties of UAP \cite{moosavi2017universal}. 
Furthermore, Mopuri et al. \cite{mopuri2017fast} introduced Fast Feature Fool as another approach to generate universal perturbations. Both transferability and generalisability of the examples were exhibited in their experiments. However, unlike the universal perturbation proposed by Moosavi-Dezfooli et al. \cite{moosavi2017universal}, Fast-Feature Fool does not require any input data to be trained on target classifiers. Hence, it is data independent. This property brings many advantages to real-world security applications where the target classifier and training data are unknown.

\subsubsection{One Pixel Attack}
Although the majority of existing attacks aim to modify all pixel values, lately Su et al. \cite{su2019onepixel} generated adversarial examples by perturbing only a single pixel. One-Pixel Attack applies differential evolution, a type of evolutionary algorithm, to perturb the colour values of a pixel and minimise the probability of the target class. Therefore, this approach can be  distinguished from other attacks, particularly UAP\cite{moosavi2017universal}, which perturbs all pixel values. Moreover, this method only requires probability values generated from the SoftMax layer of classifiers and does not need any understanding of the inner parameters. Their experiments showed generalisability  of the attack on different types of networks. 
\subsection{Adversarial Defences}
\subsubsection{Adversarial Training}
One of the first defence approaches to enhance the robustness of deep neural networks is to augment perturbed images at training stage, allowing the network to gain more knowledge about adversarial examples \cite{goodfellow2014explaining}. The state-of-the-art performance of adversarially trained models is shown in many studies \cite{madry2017towards,wang2019bilateral, moosavi2016deepfool, goodfellow2014explaining,wong2020fast,shafahi2019adversarial}.
However, the analysis of these studies is only limited to small data sets due to the expensive computational cost of training. To date, only a few studies have evaluated adversarial training at large-scale on ImageNet data set \cite{DengImageNet}, some presented network robustness against non-iterative attacks \cite{kurakin2016adversarial, tramer2017ensemble, wong2020fast}, and others on iterative adversarial examples \cite{shafahi2019adversarial, kannan2018adversarial}. 

Besides network robustness improvement, some suggest that adversarial training aims for regularisation to avoid overfitting \cite{goodfellow2014explaining,kurakin2016adversarial}, and robust optimisation, when trained with non-iterative adversary \cite{madry2017towards}. Moreover, the experiments of Wu et al. \cite {wu2017adversarialPercision} revealed that adversarial training improves precision. Madry et al. \cite{madry2017towards} also presented that PGD trained networks can be viewed as a transferable defence technique against other types of adversaries.

\subsubsection{Bayesian Defences and Probabilistic Approaches}
The essence of adversarial examples is arguably owing to the uncertainties of neural networks about predictions, as well as the universal logit distribution \cite{cubuk2017intriguing}. Following this, other types of countermeasure systems aim to defend against adversarial attacks by employing statistical techniques at the deep convolutional layers \cite{li2017adversarial,hosseini2017blocking,grosse2017statistical,abbasi2017robustness}. This is due to the fact that adversarial examples have a noticeably different data distribution as opposed to their clean images \cite{li2017adversarial,grosse2017statistical}. 

The model uncertainty and data distribution can be a reliable indicator to distinguish perturbed images from clean images. In particular, Grosse et al. \cite{grosse2017statistical} performed statistical tests to compare the distribution of perturbed images with their counterpart clean images. They observed that adding an additional class assigned to perturbed images based on the data distribution, results in detecting a significant number of adversarial examples.

Grosse et al. \cite{grosse2017statistical} further showed that their proposed technique led to a more robust deep neural network for both black-box and white-box attacks. Hosseini et al. \cite{hosseini2017blocking} posited a probabilistic approach in a similar manner. Their method adds an extra label to compute the likelihood of an image to detect an adversarial example. The classifier is then retrained with the additional label assigned to the perturbed images, which have different probabilities. This allows the classifier to effectively reject adversarial examples and correctly classify clean images at the testing stage.

In recent years, Bayesian deep learning has attracted the interest of many computer vision researchers. Unlike standard deterministic deep learning models, which only capture model confidence, a degree of uncertainty can be exhibited by Bayesian networks \cite{gal2016dropout,kendall2017uncertainties}. The uncertain nature of adversarial examples motivated Bradshaw et al. \cite{bradshaw2017adversarial} to develop a hybrid Bayesian deep convolutional network as a defence, by applying Gaussian processes to calculate probabilities. Their experiments revealed promising results in terms of network robustness. 
Abbasi and Gagné \cite{abbasi2017robustness} took a different approach based on ensemble construction. They found that misclassified labels of adversarial examples tend to be biased towards specific classes. Consequently, they proposed a mechanism to classify the images based on the majority vote. The results of their experiment showed that their method was able to predict clean images with higher confidence by discarding perturbed images using their voting mechanism. Feinman, et al. \cite{feinman2017detecting} also applied Bayesian Deep Networks by using probabilities generated from dropout layers in CNN as a priori \cite{gal2016dropout}, which successfully detected adversarial examples .  

The vulnerability of many robust defence strategies, that are proposed in the literature against adversarial attacks, is still an open question. The study of Carlini et al. \cite{carlini2017adversarial} on the efficacy of adversarial defences revealed an interesting insight. All the investigated defences with the partial exception of Monte Carlo (MC) deep neural network, as implemented in \cite{feinman2017detecting}, failed to truly detect adversarial examples. It was found that MC networks are more difficult to attack, compared to other deterministic models. This finding is also confirmed in a recent study \cite{smith2018understanding}. Smith and Gal \cite{smith2018understanding} postulate that the reason behind the current vulnerability of MC networks is the lack of the model ability to fully capture uncertain regions. It was further stated that an ensemble of Bayesian MC networks are likely to enhance their behaviour. Motivated by these insights, we propose an image set classification approach with adversarial training using Bayesian deep learning models as a defence technique against adversarial attacks.

\subsection{Image Set Classification}
Image set classification has gained a tremendous popularity in the domain of computer vision, due to its nature of handling multiple image recognition in video sequences, as well as multi-view object classification\cite{zhao2019review}. In traditional recognition tasks, a vector of pixel values for a single image is given as an input, where each vector is assigned to a class \cite{kim2007imageset}. 

Image set classification can address the drawbacks of single-shot recognition tasks arising from the scene complexity, variations in appearance \cite{shah2016iterative,nadeem2018real,hayat2015, zhao2019review}, changes in illumination, viewpoint, different camera angles, different backgrounds, and occlusions. Image set classification recognises objects from a set of images \cite{kim2007imageset}, which collectively belong to the same class.

There are generally three types of models that are proposed in the image set classification literature, (i) parametric, (ii) non-parametric, and (iii) deep learning. \textbf{Parametric methods} \cite{arandjelovic2005face} rely on probabilistic assumptions, such as Kullback–Leibler (KL) divergence, and certain statistical distributions of samples or image sets. The main shortcoming of these methods is their failure to produce robust results in the absence of weak statistical relationships. \textbf{Non-Parametric methods} \cite{nadeem2018real, shah2017efficient, wang2012covariance} adopt linear or nonlinear representations of samples rather than statistical distributions. With non-parametric methods, image sets are characterized on geometric surfaces or exemplar images. Affine Hull Image Set Distance (AHISD) \cite{cevikalp2010face}, Sparse Approximated Nearest Points (SANP) \cite{hu2012face, hu2011sparse}, Mututal Subspace Method (MSM) \cite{yamaguchi1998face}, Discriminat Cannonical Correlations (DCC) \cite{kim2007discriminative}, Manifold-Manifold Distance(MMD) \cite{wang2008manifold}, and Manifold Discriminant Analysis (MDA) \cite{wang2009manifold}, are among other types of models used for non-parametric image set classification. 

To date, only a few works in the literature have focused on the applications of image set classification using deep learning. For instance, the study of Hayat et al. \cite{hayat2015} was among the first to use a deep learning framework, named  named Adaptive Deep Network (ADNT),  for image set classification. This framework is composed of encoders and decoders. Each contains three hidden layers connected through a common hidden layer to reconstruct the input images. For both training and testing sets, Local Binary Pattern (LBP) is computed and PCA whitening is performed at the pre-processing stage. Despite the promising performance of their deep learning model for object and facial recognition, the computational complexity of their technique is high. Furthermore, their technique relies on hand-crafted LBP features for a superior performance, in contrast of using raw images.

Later, Shah et al. \cite{shah2016iterative} proposed an Iterative Deep Learning Model (IDLM) based on one convolutional layer and recurrent neural networks for object classification. IDLM was able to overcome many of the shortcomings of the previous deep learning models \cite{hayat2015}, including performance, speed, and hidden layer size, by iteratively learning feature representations. This technique achieved state-of-the-art performance on benchmark image set datasets. However, IDLM requires the tuning of several parameters.

In the view of the above, it can be noted that countermeasures against adversarial examples, however, reveal a number of research gaps. To the best of our knowledge, all current defence methods are geared towards single-shot object recognition and to date, there is no research on image set classification as a defence against adversarial attacks. 

A review of empirical results in the recent literature shows a strong robustness of deep learning applications for image set classification \cite{shah2016iterative, hayat2015}, which can be applied as a defence framework. Nevertheless, there are a number of shortcomings that have not been addressed yet in this domain. \textbf{First}, previous studies have almost exclusively applied the task of image set classification on deterministic deep learning models. \textbf{Second}, the majority of their experiments are constrained to relatively small object recognition data sets, such as ETH-80 \cite{ETH80}. \textbf{Third}, the application of modern deep learning models for image set classification remains unstudied. In this paper, we address these limitations of the existing techniques and propose a novel image set classification based approach for defence against adversarial attacks. To the best of our knowledge this is the first ever image set based defence approach against adversarial attacks.

\section{Proposed Methodology} \label{method}
\begin{figure*}
  \centering

    \includegraphics[width=1.0\textwidth]{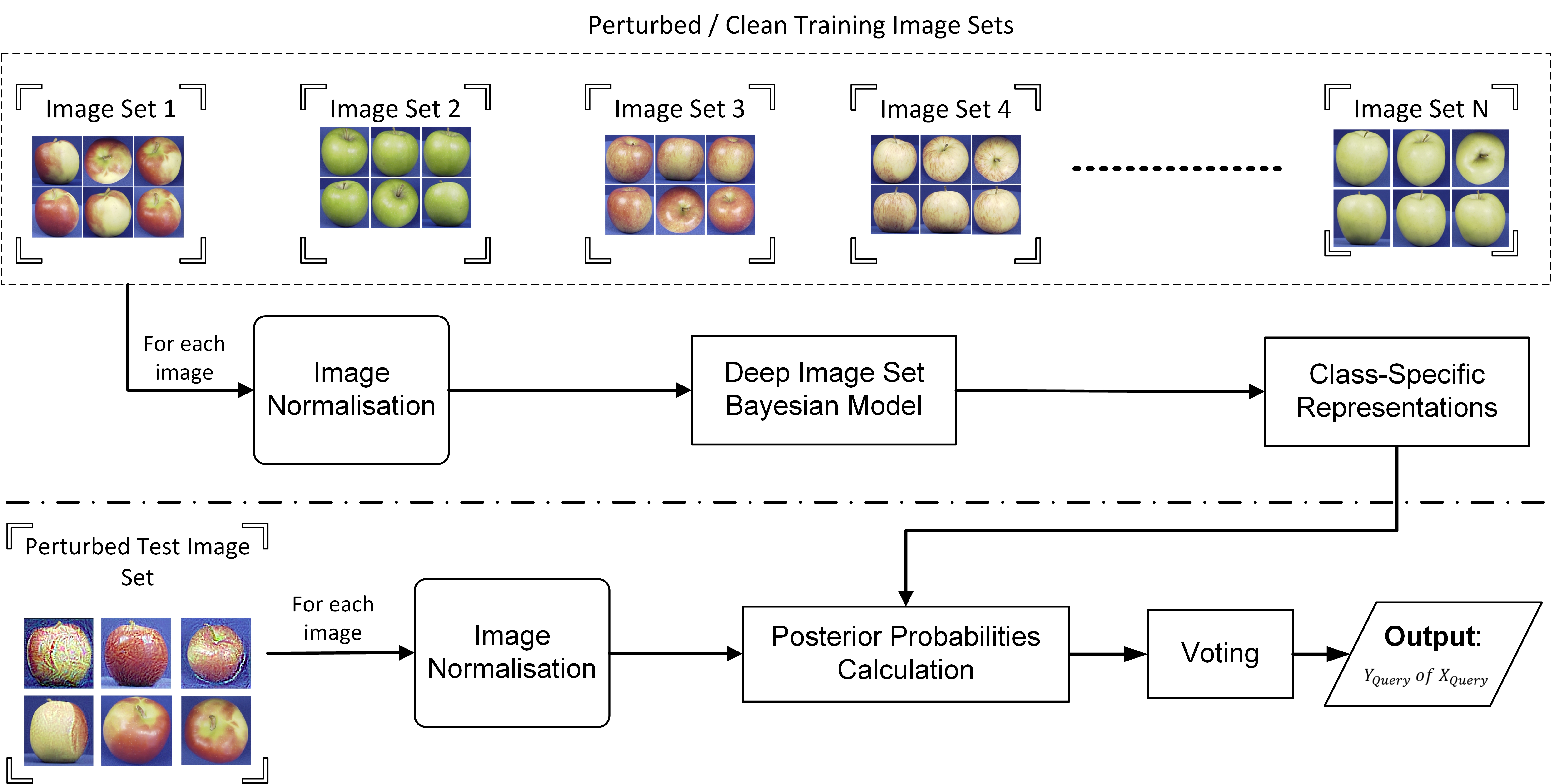}
  \caption{Block diagram of the proposed Deep Bayesian Image Set Classification Technique. The framework performs training and testing. During training, we first implement our deep image set Bayesian model, which is then trained with clean and perturbed training images of each class to learn class-speciﬁc representation. During testing, posterior Monte Carlo probabilities are computed for the image set (containing clean and perturbed images) and a voting strategy is adopted to predict the class of the test image set.}
  \label{Network Arch figure}
\end{figure*}

Figure. \ref{Network Arch figure} provides the block diagram of our proposed deep Bayesian image set classification technique for defense against adversarial attacks.

\subsection{Image Set Classification: A Defence Strategy}
In the problem of image set classification, a gallery of training images is composed of multiple image sets. Each image set contains a group of images of the same label, which has similar appearance variations. 
Let $I_{1}, I_{2}, I_{3}, ...., I_{d}$  be the $d$ images of an image set $X_{i}$, we can then define an image set as:
\begin{equation} 
\label{imageset_containingimages}
    X_{i} = \{I_{1}, I_{2}, I_{3}, ...., I_{d}\} 
\end{equation}

\begin{equation}
\label{zigma_imageset}
    G_{j} = \{X_{1}, X_{2}, ... , X_{n}\} \ \forall j \in C
\end{equation}
where $X_{1}, X_{2}, X_{3}, ...., X_{n}$ are $n$ image sets of a gallery $G_{j}$, and $C$ is the total number of classes. Equation \ref{total_gallery} defines the total number of galleries in a data set.
\begin{equation}
    \label{total_gallery}
    G_{j}^C = \sum_{j=1}^{C} G_{j}
\end{equation}
Therefore, the total number of images in each gallery $G_{j}$, equals to $n\times d$, as in Equation \ref{imageset_memberofGallery}:
\begin{equation}
\label{imageset_memberofGallery}
   \forall i \in n \ \exists \ X_{i} \in G_{j} \longrightarrow G_{j} = \{I_{1}, ..., I_{ n\times d}\} \, \forall j \in C
\end{equation}

%
%
%
%

\subsubsection{Training} 
In this section, we explain the training procedure for image set classification. If a gallery has a multiple number of training image sets, we combine all the training image sets, as one set as in Equation \ref{total_set_training}:

\begin{equation}
    \label{total_set_training}
    X_{train} = \{X_{1}, X_{2},X_{3},.... , X_{N}\}
\end{equation}
where $N$ is the total number of sets created for a dataset. We are now ready to describe our proposed adversarial training procedure. 
Given a gallery $G_{j}$ with $n$ number of image sets, the adversarial examples are generated for all image sets as $X_{i}^{'(N)}$ against the baseline classifier $f_{baseline}$. They are then augmented to model $f_{adv}$ architecture to be trained, as explained in Algorithm \ref{Adversarial_Training}. Equation \ref{adversarial_image_set} represents a training image set $X_{i}^{'}$, containing perturbed images: 

\begin{equation}
    \label{adversarial_image_set}
    X_{i}^{'} = \{I_{1}^{'}, I_{2}^{'}, I_{3}^{'}, ...., I_{d}^{'}\} \ \ \forall i \in N
\end{equation}

\begin{equation}
    \label{adversarial_image_set_all}
    X_{train}^{'} = \{X_{1}^{'}, X_{2}^{'}, ... , X_{N}^{'}\}
\end{equation}
where $X_{train}^{'}$ is the perturbed training set, which  contains all the perturbed training image sets $N$. All the images are normalised by dividing them with the maximum pixel value in each image set. Algorithm \ref{Adversarial_Training} explains our training algorithm.

\begin{algorithm}
\caption{Adversarial Training Using Image Sets}\label{Adversarial_Training}
\begin{algorithmic}[1]

\ForAll{$G_{j}$ in $G_{j}^{C}$} %
\If{$G_{j} \not\ni X_{i}$}
\State Create $n$ number of image sets $X_{i}^{n}$ for $G_{j}$
\EndIf
\EndFor
%
\State \textbf{Input 1:} Clean training image sets $X_{train}$ 
\State \textbf{Input 2:} Perturbed training image sets $X_{train}^{'}$
\State Normalise $\forall I \in X_{train}$ 
\State Normalise $\forall I^{'} \in X_{train}^{'}$ 
\State Train $f_{adv}(X_{i} \cup X_{i}^{'} )$ 
\State \textbf{Output:} $f_{adv}$
\end{algorithmic}
\end{algorithm}
%
%
%

\subsubsection{Testing} 
Let $X_{test(i)}$ be a test image set consisting of clean images $I_{t}^{d}$, and $X_{test(i)}^{'}$ be its counterpart image set containing perturbed images $I_{t}^{'d}$. We randomly perturb $R$ percent of images $I_{t}^{d}$ in $X_{test(i)}$ (Algorithm \ref{Image_set_perturb}). This produces $S_{k}$, an image set containing $n$ number of images. Algorithm \ref{Image_set_perturb} provides the procedure to perturb images sets that allows us to perform our experiments with different attack ratios $R$.
\begin{algorithm}[t]
\caption{Perturb Testing Image Set}\label{Image_set_perturb}
\begin{algorithmic}[1]
\Procedure{Perturb}{$X$}
 \State \textbf{Input:} $X_{test(i)}$, $X_{test(i)}^{'}$, $R$
    \For{$G_{test(j)}$ in $G_{test(j)}^{C}$} 
        \State $S_{1}^{R}$ = randomly select $I_{t}^{'}$ from $X_{test(i)}^{'}$ 
       \State $S_{2}^{(n-R)}$ = randomly select $I_{t}$ from $X_{test(i)}$ 
       \State $S_{k} = S_{1} \bigcup S_{2} $
\EndFor
%
\State \textbf{return} $S_{k}^n$
\EndProcedure
\end{algorithmic}
\end{algorithm}
After image normalisation, the proposed image set based defence technique (Algorithm \ref{Image_set_defence}) starts by calculating the posterior Monte Carlo probabilities $p_{s_{k}}$ of image set classifier $f_{adv}$ according to Equation \ref{mc_posterior_probability_image_set}, for each perturbed image set $S_{k} \in G_{test(j)}$.
Each $S_{k}$ casts a vote for $G_{j}$ in $y_{test}$ to calculate the accuracy. Note that $T$ is the number of stochastic forward passes as in Equation \ref{mc_posterior_probability_image}.

\begin{algorithm}
\caption{Image Set Defence}\label{Image_set_defence}
\begin{algorithmic}[1]
\Procedure{Defend}{$X$}
 \State \textbf{Input:} $S_{k}^n$, $T$
    \For{ each gallery $G_{test(j)}$ in $G_{test(j)}^{C}$} 
        
        \For{ each image set $S_{k}$ in $G_{test(j)}$}
        \State Normalise $S_{k}$
            \For{t in $T$}
                \State $p_{S_{k}} = $  Predict($S_{k}$, $f_{adv}$ ) \Comment{See Eq. (\ref{mc_posterior_probability_image}-\ref{mc_posterior_probability_image_set})}
            \EndFor
        \State MV: $\hat{y}_{S_{k}} = mode{(y_{S_{k}})}$ \Comment{See Eq. (\ref{predicted_y_image_set}-\ref{MV_hard})}
        \State SV: $\hat{y}_{S_{k}} = \arg \max(V_{bv}({S_{k}}))$ \Comment{See Eq. (\ref{soft_vote_perclass}-\ref{soft_vote_argmax})}
        \State WV: $\hat{y}_{S_{k}} = \arg \max(WV_{S_{k}})$ \Comment{See Eq. (\ref{weighted_vote_label}-\ref{y_hat_weighted})}

\EndFor
\EndFor
\State \textbf{return} Label $y_{test}$ of $X_{test}$
\EndProcedure
\end{algorithmic}
\end{algorithm}

One of the limitations of the standard deep learning frameworks is the lack of ability to accurately recognise noisy data points, which are far from training data. In such cases, the model confidence generated from the last SoftMax layer is, in fact, insufficient to determine the class of input data \cite{gal2016dropout, kendall2017uncertainties}. Hence, they can be interpreted as merely deterministic. Dropout layers applied at the training stage before every layer, in which weights are initialised, can be used as Dropout approximation\cite{gal2016dropout}. The dropout distribution at the testing stage computes variational inference by collecting Monte Carlo samples $T$ times, known as stochastic forward passes. This technique can minimise the Kullback-Leibler (KL) divergence to model the posterior probability $p$, while approximating the Gaussian Process (GP).

We apply Bayesian deep learning \cite{gal2016dropout} to approximate the ensemble of posterior distribution $p_{I_{d}}$ from the Monte Carlo distribution $M_{I_{d}}^{*}$ for an image $I_{d}$ by calculating the mean of $T$ stochastic forward passes, as shown in Equation \ref{mc_posterior_probability_image}.

\begin{equation} \label{mc_posterior_probability_image}
  p_{I_{d}} = \frac{1}{T} \sum_{t=1}^{T} f_{adv}(M_{I_{d}}^{*}) %
\end{equation}

Therefore, there are $d$ predicted probabilities for an image set $X_{test(i)}$ given by:

\begin{equation} \label{mc_posterior_probability_image_set}
  P_{X_{i}} = \{ p_{I_{1}},  p_{I_{2}}, ... ,  p_{I_{d}} \}
\end{equation}

We experimented with different voting strategies for the classification of image sets, including the exponential weighted voting, the majority voting and the soft voting, to test the performance of our proposed technique. Exponential weighted voting produced the best result for most of the data sets.
\subsubsection{Majority Vote (MV)}
Given a test image set $X_{test(i)}$ as in Equation \ref{imageset_containingimages}, each image $I$ cast an equal vote for a class $c$ to determine the predicted label $y_{x_{test(i)}}$ for an image set $x_{test(i)}$.
\begin{equation} \label{predicted_y_image}
    \hat{y}_{I_{d}} = \arg\max (p_{I_{d}})
\end{equation}

\begin{equation} \label{predicted_y_image_set}
    y_{x_{test(i)}} = \{\hat{y}_{I_{1}}, \hat{y}_{I_{2}}, ... , \hat{y}_{I_{d}}\}
\end{equation}

The most frequent label $\forall I_{d} \in X_{i}$ determines the final vote for the image set $X_{test(i)}$. 

\begin{equation} \label{MV_hard}
   \hat{y}_{x_{test(i)}} = mode{(y_{x_{test(i)}})}
\end{equation}

\subsubsection{Soft Vote (SV)}

This ensemble voting strategy determines how certain a model $f$ is in predicting the gallery label of a set $X$ at each $T$ stochastic forward pass. The posterior predicted probability of an image set $X_{test(i)}$ per each class of $c$ distributed over Monte Carlo sample $M^{*}$ is calculated as $\forall c \in C$: %

\begin{equation} \label{soft_vote_perclass}
    P_{(C_{c}|X_{test(i)})} = \frac{1}{d} \sum_{l=1}^{d} p_{(I_{d}|C_{c})} 
\end{equation}
Hence, each iteration of the stochastic forward pass $T_{t}$ contains $c$ sets of predicted probabilities given an image set $X_{test(i)}$, which belongs to class $C$ :

\begin{equation}
    T_{t} = \{P_{(C_{1}|X_{i})}, P_{(C_{2}|X_{test(i)})}, ... , P_{(C_{c}|X_{test(i)})} \} \ \forall \, t \in T 
\end{equation}

The class with the highest probability wins the vote to predict the label of an image set $X_{test(i)}$.

\begin{equation}
    V_{bv} = \max{(P_{(C_{c}|X_{test(i)}) \in \forall T_{t}})}
\end{equation}

\begin{equation} \label{soft_vote_argmax}
    \hat{y}_{x_{test(i)}} = \arg \max(V_{bv})
\end{equation}

\subsubsection{Exponential Weighted Vote (EWV)}
In this voting strategy, each image $I_{d}$ in a test image set $X_{test(i)}$ casts a vote $V_{w}$ for each class $c$ in the gallery. The exponential weighted voting for an image can be defined as: 

\begin{equation} \label{weighted_vote_label}
    V_{w(I_{d})} = e^{-\beta p_{I_{d}}}
\end{equation}
where $\beta$ is a constant. Thus, the predicted vote for the image set $X_{test(i)}$ is given by:

\begin{equation} \label{image_set_weighted}
    WV_{x_{test(i)}} = \sum_{I=1}^d V_{w(I_{d})}
\end{equation}
The class with the maximum of the accumulated weighted vote makes the final decision for the test image set:
\begin{equation} \label{y_hat_weighted}
    \hat{y}_{x_{test(i)}} = \arg \max(WV_{x_{test(i)}})
\end{equation}

%
%
%

%
%
%

%

%

\subsubsection{Proposed Deep Bayesian Model} %
The backbone architecture of our defence model is inspired by the Res-Net50 \cite{He_ResNet} architecture. To collect a sample of Monte Carlo estimates at the training and testing stage, we build on top of Res-Net50 by adding a dropout layer of 0.5 after the last activation layer and before the last fully connected layer, where the network weights are initialised. The network is then trained with the training data using fine-tuning.

\subsubsection{Baseline (VGG-16)} 
As our baseline model, we adopt the architecture of VGG-16 network \cite{simonyan2014very}, and add 3 dropout layers after the last pooling layer and before the two fully connected layers, where the network weights are initialised. This will collect a sample of Monte Carlo estimates at the testing state. The model architecture is designed with a dropout ratio of 0.5. Similar to our proposed model, fine-tuning is performed.
\section{Experimental Results} \label{experimental result}
In this section, we describe the benchmark detasets used for the evaluation of the proposed image set classification technique and single-shot image recognition (Section \ref{subsection datasets}). Our experimental results are reported in Section \ref{results}. 

\begin{table}
\renewcommand{\arraystretch}{1.3}
\caption{Performance Evaluation of the Proposed Technique on CIFAR-10 Data Set Against FGSM, PGD, and DeepFool Attacks. $\epsilon$ represents Perturbation Magnitude.}
\label{CIFar_Table_All}
\centering
\begin{tabular}{|c|p{4em}|p{2em}|c|p{2.5em}|p{2.5em}|c|c|c|c|}
\hline
Model & Attack & Adv & $\epsilon$ & Attack & SV & MV & EWV \\
      & Type   & Train &          & Ratio  &  (\%)  & (\%)   &  (\%)    \\
\hline
Baseline & FGSM   & No &    -   & 0\%     & 98.95 & 99.30 & 98.95 \\
Baseline & FGSM   & No & 0.05 & 50\%   & 96.16 & 96.16 & 96.16 \\
Baseline & FGSM   & No & 0.05 & 80\%   & 96.16 & 94.42 & 96.51 \\
Baseline & FGSM   & No & 0.05 & 100\%   & 96.51& 95.12 & 96.51 \\
Baseline & FGSM   & Yes & 0.05 & 0\%     & 98.95 & 98.60 & 98.95 \\
Baseline & FGSM   & Yes & 0.05 & 50\%   & 97.56 & 96.86 & 97.56 \\
Baseline & FGSM   & Yes & 0.05 & 80\%   & 97.90 & 97.21& 97.90 \\
Baseline & FGSM   & Yes & 0.05 & 100\%   & 98.25 & 97.56 & 98.60 \\
          &       &       &       &       &       &   &      \\
Proposed & FGSM  & No &    -   & 0\%     & \textbf{100.00} & 99.30 & \textbf{100.00} \\
Proposed & FGSM  & No & 0.05 & 50\%   & 96.86 & 96.86 & 96.86 \\
Proposed & FGSM  &  No & 0.05 & 80\%   & 87.45 & 85.71 & 87.45 \\
Proposed & FGSM  & No &0.05 & 100\%     & 74.21 & 73.51 & 74.56 \\
Proposed & FGSM  & Yes & 0.05 & 0\%     & \textbf{100.00} & \textbf{100.00} & \textbf{100.00} \\
Proposed & FGSM  &  Yes & 0.05 & 50\%   & 98.95 & 98.60& 98.95 \\
Proposed & FGSM  & Yes & 0.05 & 80\%   & \textbf{100.00} & 99.30 & \textbf{100.00} \\
Proposed & FGSM  &  Yes & 0.05 & 100\%   & \textbf{100.00} & 99.65 & \textbf{100.00} \\
\hline
Baseline & PGD   & No & - & 0\%     & 98.95& 99.30 & 98.95 \\
Baseline & PGD   & No & 0.01  & 50\%   & 98.25 & 97.21 & 97.90 \\
Baseline & PGD   & No & 0.01  & 80\%   & 98.95 & 97.56 & 98.95\\
Baseline & PGD   & No & 0.01  & 100\%     & 98.95 & 98.60 & 98.95 \\
Baseline & PGD   & Yes &- & 0     & \textbf{100.00} & 99.30 & 99.61 \\
Baseline & PGD   & Yes & 0.01  & 50\%   & 97.90 & 96.86 & 97.56 \\
Baseline & PGD   & Yes & 0.01  & 80\%   & 97.90& 97.56 & 98.60 \\
Baseline & PGD   & Yes & 0.01  & 100\%     & \textbf{100.00} & 98.95 & 99.65 \\
          &       &       &       &       &       &       &  \\
Proposed & PGD   & No &- & 0     & \textbf{100.00} & 99.30 & \textbf{100.00} \\
Proposed & PGD   & No & 0.01  & 50\%   & 98.95 & 97.56 & 98.95 \\
Proposed & PGD   & No & 0.01  & 80\%   & 99.65 & 98.95& 99.65 \\
Proposed & PGD   & No & 0.01  & 100\%     & 99.65 & 98.95& 99.65 \\
Proposed & PGD   & Yes & - & 0\%     & \textbf{100.00} & 99.60 & \textbf{100.00} \\
Proposed & PGD   & Yes & 0.01  & 50\%   & 98.25& 96.16 & 98.25 \\
Proposed & PGD   & Yes & 0.01  & 80\%   & 98.60& 98.25 & 98.60 \\
Proposed & PGD   & Yes & 0.01  & 100\%     & \textbf{100.00} & 99.60 & \textbf{100.00} \\
\hline
Baseline & DeepFool & No & - &    0\%    & 98.95 & 99.30 & 98.95 \\
Baseline & DeepFool  & No & - &    50\%   & 94.42 & 91.63 & 94.42 \\
Baseline & DeepFool  & No & - &   80\%   & 56.44 & 50.17 & 57.14 \\
Baseline & DeepFool  & No & - &    100\%     & 40.06 & 35.88 & 39.02 \\
Baseline & DeepFool  & Yes & - &  0\%     & 99.65 & 98.60& 99.65 \\
Baseline & DeepFool  & Yes & - &  50\%   & 96.86 & 94.77 & 96.16 \\
Baseline & DeepFool  & Yes & - & 80\%   & 95.12 & 91.28 & 95.81 \\
Baseline & DeepFool  & Yes & - &  100\%     & 93.37 & 88.50 & 94.45 \\
    &           &       & - &                     &           &       &         \\  
Proposed & DeepFool  & No & - & 0\%     & \textbf{100.00} & 99.30 & \textbf{100.00} \\
Proposed & DeepFool  & No & - &    50\%   & 97.90 & 90.50 & 97.90 \\
Proposed & DeepFool  & No & - &  80\%   & 47.03 & 43.90 & 47.30 \\
Proposed & DeepFool & No & - & 100\%     & 32.75 & 32.40& 32.75 \\
Proposed & DeepFool  & Yes & - &  0\%     & \textbf{100.00} & 99.30& \textbf{100.00}\\
Proposed & DeepFool  & Yes & - &  50\%   & 98.95 & 96.86 & 98.60 \\
Proposed & DeepFool  & Yes & - &  80\%   & 99.30 & 96.16 & 98.60 \\
Proposed & DeepFool  & Yes & - &  100\%     & 96.86& 93.72 & 96.86 \\
\hline
\end{tabular}%
\end{table}%
\subsection{Data Sets} \label{subsection datasets}
\subsubsection{ETH-80 Dataset}
ETH-80 dataset \cite{ETH80} has been widely used for image set classification for the task of object recognition. The dataset consists of eight object categories, namely apples, tomatoes, cows, dogs, horses, cars, and cups. Each object category is considered as gallery, which are composed of ten image sets. Each image set contains images of the object taken from 41 different viewpoints. Each image sets of a gallery has images from different types of associated objects. For instance, different breeds of horses or different types of pears. In total, there are 41 image sets, 8 galleries and 3280 images in the dataset. For our experiments, we use $128\times128$ cropped images. We randomly select 32 images from each image set for training and 9 images for testing to keep 80:20 split ratio.

\subsubsection{CIFAR-10 Dataset}
\begin{figure}
  \centering
  \includegraphics[width=0.48\textwidth]{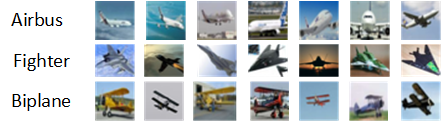}
  \caption{Examples of CIFAR-10 Image Sets for Airplane Gallery Sample. Images from three different image sets, including Airbus, Fighter, and Biplane are shown.}
  \label{CIFAR-10 images}
\end{figure}

The CIFAR-10 dataset \cite{Cifar10krizhevsky2009learning} contains 10 object classes: airplane, automobile, bird, cat, deer, dog, frog, horse, ship, and truck. There are 60,000 colour images of $32 \times 32$, among which 50,000 are for training and 10,000 for testing. Each class has 5000 images for training and 1000 images for testing. We consider each class as a gallery. We create image sets for all the galleries in the training set, based on different types of objects. For example, for the airplane gallery, all airbus, fighters or biplane images are assigned to 3 different image sets, respectively as shown in Figure. \ref{CIFAR-10 images}. Similarly, this process is repeated for all images of each gallery in the testing set. Image sets containing less than 4 images in the test set were removed. Our experiments are performed for 287 image sets belonging to 10 classes in the testing set and 9,869 images altogether.
\begin{table}
\renewcommand{\arraystretch}{1.3}
\caption{Performance Evaluation of the Proposed Technique on MNIST Data Set Against FGSM, PGD, and DeepFool Attacks. Attack ratio is set to 100\%. $\epsilon$ represents Perturbation Magnitude.}
\label{MNIST_Table_All}
\centering

\begin{tabular}{|c|p{4em}|p{2em}|c|c|c|c|p{0.5em}|p{1em}|p{0.5em}|}
\hline
Model &  Attack  & Adv  & $\epsilon$  & SV  & MV  & EWV  \\
      & Type   & Train &   & (\%) & (\%) &  (\%) \\
\hline
Baseline & FGSM  & No & 0.3      & 18.09 & 18.09 & 19.04 \\
Baseline & FGSM  & Yes & 0.3      & \textbf{100.00} & \textbf{100.00} & \textbf{100.00} \\
Proposed & FGSM  & No & 0.3        & 53.33 & 54.28 & 53.33 \\
Proposed & FGSM  & Yes & 0.3      & \textbf{100.00} & \textbf{100.00} & \textbf{100.00} \\ \hline 

Baseline & PGD    & No & 0.1        & 90.47 & 90.47 & 90.47 \\
Baseline & PGD    &   Yes & 0.1       & \textbf{100.00} & \textbf{100.00} & \textbf{100.00} \\
Proposed & PGD   & No & 0.1       & \textbf{100.00} & \textbf{100.00} & \textbf{100.00} \\
Proposed & PGD    & Yes & 0.1        & \textbf{100.00} & \textbf{100.00} & \textbf{100.00}  \\ 
\hline

Baseline & DeepFool & No & -       & 17.14  & 18.10  & 18.10 \\
Baseline & DeepFool & Yes & -       & 64.76  & 60.95  & 64.76 \\
Proposed & DeepFool & No & -      & 22.85  & 23.80  & 22.85 \\
Proposed & DeepFool & Yes & -      & 79.04  & 78.09  & 80.00 \\

\hline
\end{tabular}%
\end{table}%
\subsubsection{MNSIT}
MNIST dataset \cite{MNISTlecun1998gradient} is a collection of black and white handwritten digit images with 10 classes from zero to nine digits. The original image size in this dataset is $28 \times 28$. The dataset consists of 60,000 images for training and 10,000 images for testing. Similar to CIFAR-10, this dataset has only been used for single-image classification. Due to the similarity of images in each gallery, we assign 100 images per set. 
After this modification, we perform our experiments for 105 image sets in the testing set for the ten classes. Furthermore, the images are up-scaled to $32 \times 32$ to make them compatible with our baseline model. 
\subsubsection{Tiny ImageNet}
Tiny-ImageNet is a subset of ImageNet Large Scale Visual Recognition Challenge (ILSVRC)  \cite{russakovsky2015imagenet}. The task of object recognition and the generation of adversarial examples for this dataset is more challenging.
Hence, it can assess the generalisability of the proposed technique. The original dataset contains 100,000 training images of $64 \times 64$ resolution belonging to 200 classes. There are also 50 images per class for validation and 10,000 unlabelled images for testing.

For our experiments, we assign 10 image sets per gallery in the training set. Each set contains 50 images. All the 50 images of the validation classes are considered as one set  of their associated gallery. In addition, we randomly select two image sets from each training class in order to add to our test set. Thus, after modification, the dataset resulted in 3 image sets for each gallery of testing set, and 8 image sets per gallery for training. In summary, there are collectively 80,000 images, belonging to 1600 image sets associated with 200 galleries for training and 30,000 images for testing. The experiments are conducted for each of the 200 galleries, containing 3 sets and collectively 150 images for each class in our testing set.

\subsection{Implementation Details}
The proposed framework was implemented using TensorFlow \cite{abadi2016tensorflow} and Keras libraries available in Python. The experiments were run on a Windows machine with 16 GB RAM, 2.20 GHz Intel Xeon Silver CPU and 12 GB NVIDIA GRID T4-4Q GPU. The adversarial attacks were generated with Adversarial Robustness Toolbox \cite{nicolae2018adversarial}, developed by IBM for research purposes.

\subsection{Network Hyperparameters}
The baseline model as well as the proposed model was designed with Adam optimiser, learning rate is set to 0.0001, and batch size of 64 is used for CIFAR-10, MNIST, and Tiny-ImageNet. The batch size of 32 was selected for ETH-80, along with the same values of parameters as stated above. 

\subsection{Quantitative Results} \label{results}
We extensively evaluate the performance of our proposed image set based defence technique against a wide range of attacks, including FGSM \cite{goodfellow2014explaining}, PGD \cite{madry2017towards}, and DeepFool \cite{moosavi2016deepfool} on these datasets. In these experiments, we train our baseline model with clean images. We followed the experimental protocol of MagNet \cite{meng2017magnet}, and ME-NET \cite{yang2019meNet}, which have shown state-of-the-art performance against adversarial attacks on CIFAR-10 and TinyImageNet datasets. 

We then randomly perturb the images in each image set based on different fractions $R$, as in Algorithm \ref{Image_set_perturb}, including 50\%, 80\%, and 100\%, to conduct our experiments in both white-box and black-box settings. The experiments are performed for 50 stochastic forward passes ($T=50$). 

\subsubsection{Performance on CIFAR-10}
Table \ref{CIFar_Table_All} reports the performance of our proposed technique against FGSM, PGD, and DeepFool attacks. Fast Gradient Method (FGSM) attack is not able to considerably fool the network in white-box setting. However, it fools the unseen proposed network and decreases the classification accuracy on clean images by 26\%. These results confirm the transferability of FGSM and DeepFool. Adversarial training improves the robustness of both networks when all the images of the testing sets are perturbed. We observe that imperceptible adversarial examples, with $\epsilon = 0.05$, does not noticeably fool the proposed baseline VGG-16, compared to Res-Net50. 
Similarly, Table \ref{CIFar_Table_All} shows a strong robustness of our network against PGD. In constrast, DeepFool attack significantly fools the proposed baseline VGG-16 classifier whithout adversarial training by virtually 58\% in white-box setting. Adversarial training of VGG-16 classifier and applying image set classification demonstrates the robustness of the defence model. Moreover, we also tested our  proposed technique in a black-box setting. DeepFool has adverse effect on fooling the clean model, while the defence exhibited a strong robustness. 

\begin{table}
\renewcommand{\arraystretch}{1.3}
\caption{Performance Evaluation of the Proposed Technique on Tiny ImageNet Data Set Against FGSM, PGD, and DeepFool Attacks. Attack ratio is set to 100\%. $\epsilon$ represents Perturbation Magnitude.}
\label{Tiny_ImageNet_Table_All}
\centering
\begin{tabular}{|c|p{4em}|p{2em}|c|c|c|c|p{0.5em}|p{1em}|p{0.5em}|}
\hline
Model & Attack  & Adv & $\epsilon$  & SV & MV & EWV \\
      & Type    & Train &        & (\%) &(\%) & (\%) \\         
\hline
Baseline & FGSM  & No & 0.3     & 1.16 & 1.50 & 1.00 \\
Baseline & FGSM  & Yes & 0.3    & 61.16 & 50.50 & 61.50 \\
Proposed & FGSM  & No & 0.3     & 0.66 & 0.66 & 0.66 \\
Proposed & FGSM  & Yes & 0.3     & 90.16 & 84.50 & 90.16 \\
\hline
Baseline & PGD   &  No & 0.03       & 64.66 & 57.33 & 61.83\\
Baseline & PGD   &  Yes & 0.03      & 90.16 & 85.16 & 89.50 \\
Proposed & PGD   & No & 0.03     & 97.00 & 96.33 & 96.33 \\
Proposed & PGD   &  Yes & 0.03      & 99.00& 98.00 & 98.50 \\
\hline
Baseline & DeepFool & No    & -     & 86.66 & 79.50 & 86.16 \\
Baseline & DeepFool & Yes   & -      & 91.33 & 87.17 & 91.33 \\
Proposed & DeepFool & No    & -     & 97.66 & 96.00 & 98.16 \\
Proposed & DeepFool & Yes   & -     & \textbf{99.50} & 98.66 & 99.16 \\

\hline
\end{tabular}%
\end{table}%
\subsubsection{ Performance on MNIST}
Empirical tests on MNSIT dataset show that imperceptible adversarial examples do not fool the network, due to the simplicity of the dataset and depth of our networks. Therefore, we tested with a higher $\epsilon$ value of 0.3 for FGSM. FGSM attack highly fooled the baseline model as reported in Table \ref{MNIST_Table_All}, whereas it had less effect on ResNet-50. It is clear that the proposed technique significantly improved the model robustness in all white-box and black-box settings. In comparison with other attacks, DeepFool has a higher fooling rate. It can be noted that the adversarial training of both models improves the performance of both models. 
In Section \ref{ablative} (Ablative study), we analyse the effect of attack ratio i.e., the number of the perturbed images on the performance of our model.

\subsubsection{Performance on Tiny-ImageNet}
Table \ref{Tiny_ImageNet_Table_All} reports the performance of our proposed technique on Tiny-ImageNet dataset against PGD and FGSM attacks, respectively. It can be noted that applying image set classification along with deep learning increases the model performance (in terms of accuracy). Moreover, as the attack ratio increases, the FGSM attack noticeably fools the baseline network in white-box setting, unlike PGD. Regardless of the extent of clean models being fooled, the results show that adversarial training with image set classification allows deep learning models to achieve great robustness on larger datasets, which consists of several galleries. The proposed image set based defence technique provides a high degree of robustness against DeepFool attack on ResNet network without adversarial training, as reported in Table \ref{Tiny_ImageNet_Table_All}. A similar behaviour is also observed for PGD. Empirically, it was found that the underlying reason behind this behaviour lies in two factors. \textbf{First}, the variability of classes at the SoftMax layer makes the Bayesian network to be less confident about its decision on perturbed images at each stochastic forward pass. This allows the voting strategy to determine the predicted gallery of an image set based on the most certain predicted posterior probabilities of images. \textbf{Second}, adversarial examples generated for DeepFool and PGD ($\epsilon = 0.03$) are comparatively imperceptible compared to FGSM attack ($\epsilon = 0.3$) in the experiments. As a result, the proposed defence technique can make a more accurate decision by capturing the most certain images.

\subsubsection{Performance on ETH-80}
We also analyse the performance of our proposed technique on ETH-80 dataset, which has extensively been used for image set classification in the literature. Table \ref{ETH_Table_All} reports our results with a relatively high perturbation magnitude of $\epsilon = 0.3$ for PGD and FGSM, respectively. The proposed technique achieves a robust performance against these attacks as well as DeepFool as shown in Table \ref{ETH_Table_All}. These results demonstrate the robustness of the proposed image set based defence framework regardless of the adverse effects of attacks (on viewpoints). The analysis of our technique on ETH-80 casts a new light on both the image size and the type of dataset. It is observed that the technique performs well even with a considerably larger image size of $128 \times 128$, compared to other datasets. Another promising finding is that the proposed framework can also achieve a great performance regardless of the recognition tasks the input datasets are originally created for.

\subsection{Comparison with State-of-the-art} \label{StateOfTheArt}
We compare the performance of our technique with the state-of-the-art defence systems such as, MagNet \cite{meng2017magnet} and MeNet \cite{yang2019meNet} on CIFAR-10, MNIST, and Tiny-ImageNet. These technqiues have shown a superior performance against adversarial attacks, as reported in \cite{meng2017magnet} and \cite{yang2019meNet}. The results provide comparison with techniques that use either the same perturbation magnitude or lower in our experiments. Table \ref{comparison} depicts that image set classification with adversarial training outperforms other defence methods, which rely on single-shot recognition. Comparatively, the results confirm the robustness of classifying images from an image set even when collectively all the images are perturbed with adversarial examples.

\subsection{Ablative Study} \label{ablative}
We explore the effect of various parameters on the performance of the proposed defence method. In this ablative analysis, the attack ratio, the perturbation magnitude$(\epsilon)$ and the effect of image size are examined along with a comparison of the single-shot classification performance with the proposed technique. The experiments are performed on CIFAR-10 dataset.

%
\begin{table}
\renewcommand{\arraystretch}{1.3}
\caption{Performance Evaluation of the Proposed Technique on ETH-80 Data Set Against FGSM, PGD, and DeepFool Attacks. Attack ratio is set to 100\%. $\epsilon$ represents Perturbation Magnitude. Split Ratio is set to 80:20.}
\label{ETH_Table_All}
\centering
\begin{tabular}{|c|p{4em}|p{2em}|c|c|c|c|p{0.5em}|p{1em}|p{0.5em}|}
\hline
Model &  Attack  & Adv  & $\epsilon$  & SV  & MV  & EWV  \\
      & Type   & Train &   & (\%) & (\%) &  (\%) \\
\hline
Baseline & FGSM   & No & 0.3    & 25.00 & 25.00 & 25.00 \\
Baseline & FGSM   & Yes & 0.3   & 81.25 & 83.75 & 85.00 \\
Proposed & FGSM   & No & 0.3    & 12.50& 12.50 & 12.50 \\
Proposed & FGSM   & Yes & 0.3   & 78.75 & 78.75 & 80.00 \\
\hline
Baseline & PGD    & No & 0.3        & 3.75 & 3.75 & 3.75 \\
Baseline & PGD    & Yes & 0.3        & 88.75 & 85.00 & 86.25 \\
Proposed & PGD    & No & 0.3        & 12.50 & 12.50 & 12.50 \\
Proposed & PGD    & Yes & 0.3       & 72.50 & 72.50 & 73.75 \\
\hline
Baseline & DeepFool  & No & -     & 26.25 & 22.50 & 23.75 \\
Baseline & DeepFool  & Yes & -     & 88.75 & 92.50 & 91.25 \\
Proposed & DeepFool  & No & -     & 56.25 & 55.00 & 60.00 \\
Proposed & DeepFool  & Yes & -     & 93.75 & \textbf{95.00} & \textbf{95.00} \\
\hline

\end{tabular}%
\end{table}%
\subsubsection{Attack Ratio}
The analysis of different attack ratios on each image set for CIFAR-10 is shown in Table \ref{CIFar_Table_All} (column 5). It is clear that as the number of perturbed images in each image set increases, the accuracy of adversarially trained model decreases for DeepFool attack. The number of clean images in each image set leads the model to be more certain about the prediction. In contrast, even when all the image sets are perturbed, the defence model can still accurately predict the gallery labels. 
\begin{table}
\renewcommand{\arraystretch}{1.3}
\caption{Comparison with State-of-the-art Techniques. $\epsilon$ represents Perturbation Magnitude.}
\label{comparison}
\centering
\begin{tabular}{|c|c|c|c|c|c|}
\hline
Dataset         & Attack &    $\epsilon$    &   Ours  & MagNet  & MeNet  \\
         &  &        &     & \cite{meng2017magnet}  &  \cite{yang2019meNet} \\
\hline
CIFAR-10        &  FGSM     &     0.05    & \textbf{100\%}   &   \textbf{100\%}   &   92.20\% \\
CIFAR-10        &  DeepFool &       -     &   \textbf{96.86\%} & 93.40\% & - \\
CIFAR-10        &   PGD     &     0.01  &      \textbf{100\%} & -     & 91.80\% \\
MNIST           &   FGSM    &       0.3   &      \textbf{100\%} & \textbf{100\%} & - \\
MNIST           &   DeepFool &      -    &      80\%  & \textbf{99.40\%} & - \\
Tiny-ImageNet   &   FGSM    &       0.3           &  \textbf{90.16\%} & -     & 67.10\% \\
Tiny-ImageNet   &   PGD     &        0.03          &  \textbf{99\%}  & -     & 66.30\% \\

\hline
\end{tabular}%
\end{table}
\subsubsection{Perturbation Magnitude}
We tested the performance of the proposed technique on several higher perceptible perturbation magnitudes, including $\epsilon$ = 0.3, 0.5, and 0.7, with FGSM attack on CIFAR-10 dataset. Our results are reported in Table \ref{CIFAR_FGSM_03}. As can be noted, there is little to no effect of $\epsilon$ on the performance of the proposed technique. Even for $\epsilon = 0.7$, the proposed technique achieves 99.30\%, 99.30\%, 99.65\% accuracy (with adversarial training) using soft vote, majority vote and exponential voting, respectively. The reported performance on high perturbation magnitude demonstrates the generalisability of the proposed defence framework. 
\begin{table}
\renewcommand{\arraystretch}{1.3}
\caption{The Effect of Perturbation Magnitude on CIFAR-10 Data Set against FGSM Attack. $\epsilon$ represents Perturbation Magnitude. Attack Ratio is to 100\%}
\label{CIFAR_FGSM_03}
\centering
\begin{tabular}{|c|c|c|c|c|c|}
\hline
S. No. & $\epsilon$  & SV & MV & EWV \\
       &            & (\%) & (\%) & (\%) \\
\hline
 1.   & 0.05    & \textbf{100.00} & 99.65 & \textbf{100.00} \\
 2.   & 0.3        & 97.21 & 95.12 & 96.86 \\
 3.   & 0.5   & 96.16 & 93.33 & 96.16 \\
 4.   & 0.7   & 99.30 & 99.30 & 99.65\\
\hline
\end{tabular}
\end{table}
\begin{table}
\renewcommand{\arraystretch}{1.3}
\caption{The Effect of Image Size on the Proposed Technique using CIFAR-10 and MNIST datasets. Adversarial training was performed in these experiments and attack ratio is 100\%.}
\label{image_size}
\centering
\begin{tabular}{|c|c|c|c|c|}
\hline
Dataset &  Image Size & SV    & MV    & WV \\
        &             & (\%)  & (\%)  & (\%) \\
\hline
CIFAR-10    & $20\times 20$ & 97.90 & 97.90 & 97.90 \\
CIFAR-10     & $28\times 28$ &  96.86 & 96.16 & 97.21  \\
CIFAR-10     & $32\times 32$ & 96.86 & 93.72 & 96.86   \\
CIFAR-10    & $40\times40$ & \textbf{98.60} & 97.90 & 98.25 \\
\hline
MNIST     & $20\times 20$ & \textbf{100} & \textbf{100} & \textbf{100} \\
MNIST    & $28\times 28$ & \textbf{100} & \textbf{100} & \textbf{100} \\
MNIST    & $32\times 32$ &  79.04 & 78.09 & 80.00   \\
MNIST    & $40\times40$ & \textbf{100} & \textbf{100} & \textbf{100} \\
\hline
\end{tabular}%
\end{table}
\subsubsection{The Effect of Image Size}
The unexpected behaviour of the defence technique against DeepFool attack on MNIST dataset, motivated us to further analyse the effect of image size on CIFAR-10 and MNIST datasets as reported in Table \ref{image_size}. We observe that the performance of our proposed image based defence technique does not demonstrate a significant difference on smaller or larger image sizes compared to the original image size of the dataset.

We also observed that the proposed defence technique was not able to perform well specifically on MNIST, as the number of DeepFool perturbed images in each image set are increased. This is particularly owing to up-scaled image size of the dataset to make them compatible with our baseline model. We tested with the original image size of $28 \times 28$ on ResNet-50. 
The defence method performed with 100\% accuracy on the proposed network trained with adversarial examples in all voting strategies. 

\subsubsection{Single-shot classification vs. Image Set classification}
We also compare the performance of our proposed image set classification defence for FGSM attack (as presented in Table \ref{CIFar_Table_All}) with single-shot task on the Res-Net 50. The model achieves an accuracy of 74.78\% for $T = 50$ on adversarial examples in test set, whereas 100\% accuracy with the proposed image set classification defence strategy. This suggests the suitability of image set classification as a defence mechanism against adversarial attacks. 

\section{Conclusion} \label{conclusion}
We proposed image set classification as a robust defence technique against a wide range of adversarial attacks. With the help of image set classification deep learning models can gain extra knowledge on the underlying patterns of each class. We analysed the proposed technique on various datasets used for singe-shot classification along with image set classification. The proposed image set classification technique achieves the state-of-the-art performance on benchmark datasets against iterative and non-iterative adversarial attacks. We also showed that the proposed image set classification technique can be applied on any object recognition tasks by modifying the training data. The proposed technique can be used as a potential defence mechanism against adversarial examples in the physical world. 



%

\ifCLASSOPTIONcompsoc
  \section*{Acknowledgments}
\else
  \section*{Acknowledgment}
\fi
This research has been supported by Murdoch Univeristy, and is partially supported by the Australian Research Council https://www.arc.gov.au/ (Grants DP150100294 and DP150104251).

\ifCLASSOPTIONcaptionsoff
  \newpage
\fi



%

\bibliographystyle{IEEEtran}
\bibliography{references}

\begin{thebibliography}{10}
\providecommand{\url}[1]{#1}
\csname url@samestyle\endcsname
\providecommand{\newblock}{\relax}
\providecommand{\bibinfo}[2]{#2}
\providecommand{\BIBentrySTDinterwordspacing}{\spaceskip=0pt\relax}
\providecommand{\BIBentryALTinterwordstretchfactor}{4}
\providecommand{\BIBentryALTinterwordspacing}{\spaceskip=\fontdimen2\font plus
\BIBentryALTinterwordstretchfactor\fontdimen3\font minus
  \fontdimen4\font\relax}
\providecommand{\BIBforeignlanguage}[2]{{%
\expandafter\ifx\csname l@#1\endcsname\relax
\typeout{** WARNING: IEEEtran.bst: No hyphenation pattern has been}%
\typeout{** loaded for the language `#1'. Using the pattern for}%
\typeout{** the default language instead.}%
\else
\language=\csname l@#1\endcsname
\fi
#2}}
\providecommand{\BIBdecl}{\relax}
\BIBdecl

\bibitem{Lecun2015}
Y.~LeCun, Y.~Bengio, and G.~Hinton, ``Deep learning,'' \emph{nature}, vol. 521,
  no. 7553, pp. 436--444, 2015.

\bibitem{krizhevsky2012imagenet}
A.~Krizhevsky, I.~Sutskever, and G.~E. Hinton, ``Imagenet classification with
  deep convolutional neural networks,'' in \emph{Advances in neural information
  processing systems}, 2012, pp. 1097--1105.

\bibitem{chen_semantic}
L.~{Chen}, G.~{Papandreou}, I.~{Kokkinos}, K.~{Murphy}, and A.~L. {Yuille},
  ``Deeplab: Semantic image segmentation with deep convolutional nets, atrous
  convolution, and fully connected crfs,'' \emph{IEEE Transactions on Pattern
  Analysis and Machine Intelligence}, vol.~40, no.~4, pp. 834--848, April 2018.

\bibitem{He_ResNet}
K.~{He}, X.~{Zhang}, S.~{Ren}, and J.~{Sun}, ``Deep residual learning for image
  recognition,'' in \emph{2016 IEEE Conference on Computer Vision and Pattern
  Recognition (CVPR)}, 2016, pp. 770--778.

\bibitem{akhtar2018threat}
N.~Akhtar and A.~Mian, ``Threat of adversarial attacks on deep learning in
  computer vision: A survey,'' \emph{IEEE Access}, vol.~6, pp.
  14\,410--14\,430, 2018.

\bibitem{giusti2015machine}
A.~Giusti, J.~Guzzi, D.~C. Cire{\c{s}}an, F.-L. He, J.~P. Rodr{\'\i}guez,
  F.~Fontana, M.~Faessler, C.~Forster, J.~Schmidhuber, G.~Di~Caro
  \emph{et~al.}, ``A machine learning approach to visual perception of forest
  trails for mobile robots,'' \emph{IEEE Robotics and Automation Letters},
  vol.~1, no.~2, pp. 661--667, 2015.

\bibitem{esteva2017dermatologist}
A.~Esteva, B.~Kuprel, R.~A. Novoa, J.~Ko, S.~M. Swetter, H.~M. Blau, and
  S.~Thrun, ``Dermatologist-level classification of skin cancer with deep
  neural networks,'' \emph{nature}, vol. 542, no. 7639, pp. 115--118, 2017.

\bibitem{szegedy2013intriguing}
C.~Szegedy, W.~Zaremba, I.~Sutskever, J.~Bruna, D.~Erhan, I.~Goodfellow, and
  R.~Fergus, ``Intriguing properties of neural networks,'' \emph{arXiv preprint
  arXiv:1312.6199}, 2013.

\bibitem{moosavi2016deepfool}
S.-M. Moosavi-Dezfooli, A.~Fawzi, and P.~Frossard, ``Deepfool: a simple and
  accurate method to fool deep neural networks,'' in \emph{Proceedings of the
  IEEE conference on computer vision and pattern recognition}, 2016, pp.
  2574--2582.

\bibitem{madry2017towards}
A.~Madry, A.~Makelov, L.~Schmidt, D.~Tsipras, and A.~Vladu, ``Towards deep
  learning models resistant to adversarial attacks,'' \emph{arXiv preprint
  arXiv:1706.06083}, 2017.

\bibitem{su2019onepixel}
J.~Su, D.~V. Vargas, and K.~Sakurai, ``One pixel attack for fooling deep neural
  networks,'' \emph{IEEE Transactions on Evolutionary Computation}, vol.~23,
  no.~5, pp. 828--841, 2019.

\bibitem{moosavi2017universal}
S.-M. Moosavi-Dezfooli, A.~Fawzi, O.~Fawzi, and P.~Frossard, ``Universal
  adversarial perturbations,'' in \emph{Proceedings of the IEEE conference on
  computer vision and pattern recognition}, 2017, pp. 1765--1773.

\bibitem{goodfellow2014explaining}
I.~J. Goodfellow, J.~Shlens, and C.~Szegedy, ``Explaining and harnessing
  adversarial examples,'' \emph{arXiv preprint arXiv:1412.6572}, 2014.

\bibitem{mopuri2017fast}
K.~R. Mopuri, U.~Garg, and R.~V. Babu, ``Fast feature fool: A data independent
  approach to universal adversarial perturbations,'' \emph{arXiv preprint
  arXiv:1707.05572}, 2017.

\bibitem{athalye2018synthesizing}
A.~Athalye, L.~Engstrom, A.~Ilyas, and K.~Kwok, ``Synthesizing robust
  adversarial examples,'' in \emph{International conference on machine
  learning}.\hskip 1em plus 0.5em minus 0.4em\relax PMLR, 2018, pp. 284--293.

\bibitem{roadsignEykholt}
K.~{Eykholt}, I.~{Evtimov}, E.~{Fernandes}, B.~{Li}, A.~{Rahmati}, C.~{Xiao},
  A.~{Prakash}, T.~{Kohno}, and D.~{Song}, ``Robust physical-world attacks on
  deep learning visual classification,'' in \emph{2018 IEEE/CVF Conference on
  Computer Vision and Pattern Recognition}, 2018, pp. 1625--1634.

\bibitem{RoboticvisionMelis2017}
M.~Melis, A.~Demontis, B.~Biggio, G.~Brown, G.~Fumera, and F.~Roli, ``Is deep
  learning safe for robot vision? adversarial examples against the icub
  humanoid,'' in \emph{Proceedings of the IEEE International Conference on
  Computer Vision Workshops}, 2017, pp. 751--759.

\bibitem{finlayson2018adversarial}
S.~G. Finlayson, H.~W. Chung, I.~S. Kohane, and A.~L. Beam, ``Adversarial
  attacks against medical deep learning systems,'' \emph{arXiv preprint
  arXiv:1804.05296}, 2018.

\bibitem{papernot2017cyberspace}
N.~Papernot, P.~McDaniel, I.~Goodfellow, S.~Jha, Z.~B. Celik, and A.~Swami,
  ``Practical black-box attacks against machine learning,'' in
  \emph{Proceedings of the 2017 ACM on Asia conference on computer and
  communications security}, 2017, pp. 506--519.

\bibitem{kurakin2016adversarial_phone}
A.~Kurakin, I.~Goodfellow, and S.~Bengio, ``Adversarial examples in the
  physical world,'' \emph{arXiv preprint arXiv:1607.02533}, 2016.

\bibitem{wang2019bilateral}
J.~Wang and H.~Zhang, ``Bilateral adversarial training: Towards fast training
  of more robust models against adversarial attacks,'' in \emph{Proceedings of
  the IEEE International Conference on Computer Vision}, 2019, pp. 6629--6638.

\bibitem{shafahi2019adversarial}
A.~Shafahi, M.~Najibi, M.~A. Ghiasi, Z.~Xu, J.~Dickerson, C.~Studer, L.~S.
  Davis, G.~Taylor, and T.~Goldstein, ``Adversarial training for free!'' in
  \emph{Advances in Neural Information Processing Systems}, 2019, pp.
  3358--3369.

\bibitem{wong2020fast}
E.~Wong, L.~Rice, and J.~Z. Kolter, ``Fast is better than free: Revisiting
  adversarial training,'' \emph{arXiv preprint arXiv:2001.03994}, 2020.

\bibitem{tramer2017ensemble}
F.~Tram{\`e}r, A.~Kurakin, N.~Papernot, I.~Goodfellow, D.~Boneh, and
  P.~McDaniel, ``Ensemble adversarial training: Attacks and defenses,''
  \emph{arXiv preprint arXiv:1705.07204}, 2017.

\bibitem{feinman2017detecting}
R.~Feinman, R.~R. Curtin, S.~Shintre, and A.~B. Gardner, ``Detecting
  adversarial samples from artifacts,'' \emph{arXiv preprint arXiv:1703.00410},
  2017.

\bibitem{li2017adversarial}
X.~Li and F.~Li, ``Adversarial examples detection in deep networks with
  convolutional filter statistics,'' in \emph{Proceedings of the IEEE
  International Conference on Computer Vision}, 2017, pp. 5764--5772.

\bibitem{grosse2017statistical}
K.~Grosse, P.~Manoharan, N.~Papernot, M.~Backes, and P.~McDaniel, ``On the
  (statistical) detection of adversarial examples,'' \emph{arXiv preprint
  arXiv:1702.06280}, 2017.

\bibitem{Papernot_Dissiltation}
N.~{Papernot}, P.~{McDaniel}, X.~{Wu}, S.~{Jha}, and A.~{Swami}, ``Distillation
  as a defense to adversarial perturbations against deep neural networks,'' in
  \emph{2016 IEEE Symposium on Security and Privacy (SP)}, 2016, pp. 582--597.

\bibitem{carlini2017adversarial}
N.~Carlini and D.~Wagner, ``Adversarial examples are not easily detected:
  Bypassing ten detection methods,'' in \emph{Proceedings of the 10th ACM
  Workshop on Artificial Intelligence and Security}, 2017, pp. 3--14.

\bibitem{zhao2019review}
Z.-Q. Zhao, S.-T. Xu, D.~Liu, W.-D. Tian, and Z.-D. Jiang, ``A review of image
  set classification,'' \emph{Neurocomputing}, vol. 335, pp. 251--260, 2019.

\bibitem{hayat2015}
M.~{Hayat}, M.~{Bennamoun}, and S.~{An}, ``Deep reconstruction models for image
  set classification,'' \emph{IEEE Transactions on Pattern Analysis and Machine
  Intelligence}, vol.~37, no.~4, pp. 713--727, 2015.

\bibitem{shah2016iterative}
S.~A.~A. Shah, M.~Bennamoun, and F.~Boussaid, ``Iterative deep learning for
  image set based face and object recognition,'' \emph{Neurocomputing}, vol.
  174, pp. 866--874, 2016.

\bibitem{nadeem2018real}
U.~Nadeem, S.~A.~A. Shah, M.~Bennamoun, R.~Togneri, and F.~Sohel, ``Real time
  surveillance for low resolution and limited-data scenarios: an image set
  classification approach,'' \emph{arXiv preprint arXiv:1803.09470}, 2018.

\bibitem{YuanAdversarialExamples2019}
X.~{Yuan}, P.~{He}, Q.~{Zhu}, and X.~{Li}, ``Adversarial examples: Attacks and
  defenses for deep learning,'' \emph{IEEE Transactions on Neural Networks and
  Learning Systems}, vol.~30, no.~9, pp. 2805--2824, 2019.

\bibitem{zhang2019adversarial}
J.~Zhang and C.~Li, ``Adversarial examples: Opportunities and challenges,''
  \emph{IEEE transactions on neural networks and learning systems}, 2019.

\bibitem{liu2016delving}
Y.~Liu, X.~Chen, C.~Liu, and D.~Song, ``Delving into transferable adversarial
  examples and black-box attacks,'' \emph{arXiv preprint arXiv:1611.02770},
  2016.

\bibitem{tramer2017space}
F.~Tram{\`e}r, N.~Papernot, I.~Goodfellow, D.~Boneh, and P.~McDaniel, ``The
  space of transferable adversarial examples,'' \emph{arXiv preprint
  arXiv:1704.03453}, 2017.

\bibitem{jia2014caffe}
Y.~Jia, E.~Shelhamer, J.~Donahue, S.~Karayev, J.~Long, R.~Girshick,
  S.~Guadarrama, and T.~Darrell, ``Caffe: Convolutional architecture for fast
  feature embedding,'' in \emph{Proceedings of the 22nd ACM international
  conference on Multimedia}, 2014, pp. 675--678.

\bibitem{simonyan2014very}
K.~Simonyan and A.~Zisserman, ``Very deep convolutional networks for
  large-scale image recognition,'' \emph{arXiv preprint arXiv:1409.1556}, 2014.

\bibitem{DengImageNet}
J.~{Deng}, W.~{Dong}, R.~{Socher}, L.~{Li}, {Kai Li}, and {Li Fei-Fei},
  ``Imagenet: A large-scale hierarchical image database,'' in \emph{2009 IEEE
  Conference on Computer Vision and Pattern Recognition}, 2009, pp. 248--255.

\bibitem{kurakin2016adversarial}
A.~Kurakin, I.~Goodfellow, and S.~Bengio, ``Adversarial machine learning at
  scale,'' \emph{arXiv preprint arXiv:1611.01236}, 2016.

\bibitem{kannan2018adversarial}
H.~Kannan, A.~Kurakin, and I.~Goodfellow, ``Adversarial logit pairing,''
  \emph{arXiv preprint arXiv:1803.06373}, 2018.

\bibitem{wu2017adversarialPercision}
Y.~Wu, D.~Bamman, and S.~Russell, ``Adversarial training for relation
  extraction,'' in \emph{Proceedings of the 2017 Conference on Empirical
  Methods in Natural Language Processing}, 2017, pp. 1778--1783.

\bibitem{cubuk2017intriguing}
E.~D. Cubuk, B.~Zoph, S.~S. Schoenholz, and Q.~V. Le, ``Intriguing properties
  of adversarial examples,'' \emph{arXiv preprint arXiv:1711.02846}, 2017.

\bibitem{hosseini2017blocking}
H.~Hosseini, Y.~Chen, S.~Kannan, B.~Zhang, and R.~Poovendran, ``Blocking
  transferability of adversarial examples in black-box learning systems,''
  \emph{arXiv preprint arXiv:1703.04318}, 2017.

\bibitem{abbasi2017robustness}
M.~Abbasi and C.~Gagn{\'e}, ``Robustness to adversarial examples through an
  ensemble of specialists,'' \emph{arXiv preprint arXiv:1702.06856}, 2017.

\bibitem{gal2016dropout}
Y.~Gal and Z.~Ghahramani, ``Dropout as a bayesian approximation: Representing
  model uncertainty in deep learning,'' in \emph{international conference on
  machine learning}, 2016, pp. 1050--1059.

\bibitem{kendall2017uncertainties}
A.~Kendall and Y.~Gal, ``What uncertainties do we need in bayesian deep
  learning for computer vision?'' in \emph{Advances in neural information
  processing systems}, 2017, pp. 5574--5584.

\bibitem{bradshaw2017adversarial}
J.~Bradshaw, A.~G. d.~G. Matthews, and Z.~Ghahramani, ``Adversarial examples,
  uncertainty, and transfer testing robustness in gaussian process hybrid deep
  networks,'' \emph{arXiv preprint arXiv:1707.02476}, 2017.

\bibitem{smith2018understanding}
L.~Smith and Y.~Gal, ``Understanding measures of uncertainty for adversarial
  example detection,'' \emph{arXiv preprint arXiv:1803.08533}, 2018.

\bibitem{kim2007imageset}
T.~{Kim}, J.~{Kittler}, and R.~{Cipolla}, ``Discriminative learning and
  recognition of image set classes using canonical correlations,'' \emph{IEEE
  Transactions on Pattern Analysis and Machine Intelligence}, vol.~29, no.~6,
  pp. 1005--1018, 2007.

\bibitem{arandjelovic2005face}
O.~Arandjelovic, G.~Shakhnarovich, J.~Fisher, R.~Cipolla, and T.~Darrell,
  ``Face recognition with image sets using manifold density divergence,'' in
  \emph{2005 IEEE Computer Society Conference on Computer Vision and Pattern
  Recognition (CVPR'05)}, vol.~1.\hskip 1em plus 0.5em minus 0.4em\relax IEEE,
  2005, pp. 581--588.

\bibitem{shah2017efficient}
S.~A. Shah, U.~Nadeem, M.~Bennamoun, F.~Sohel, and R.~Togneri, ``Efficient
  image set classification using linear regression based image
  reconstruction,'' in \emph{Proceedings of the IEEE conference on computer
  vision and pattern recognition workshops}, 2017, pp. 99--108.

\bibitem{wang2012covariance}
R.~Wang, H.~Guo, L.~S. Davis, and Q.~Dai, ``Covariance discriminative learning:
  A natural and efficient approach to image set classification,'' in \emph{2012
  IEEE Conference on Computer Vision and Pattern Recognition}.\hskip 1em plus
  0.5em minus 0.4em\relax IEEE, 2012, pp. 2496--2503.

\bibitem{cevikalp2010face}
H.~Cevikalp and B.~Triggs, ``Face recognition based on image sets,'' in
  \emph{2010 IEEE Computer Society Conference on Computer Vision and Pattern
  Recognition}.\hskip 1em plus 0.5em minus 0.4em\relax IEEE, 2010, pp.
  2567--2573.

\bibitem{hu2012face}
Y.~Hu, A.~S. Mian, and R.~Owens, ``Face recognition using sparse approximated
  nearest points between image sets,'' \emph{IEEE transactions on pattern
  analysis and machine intelligence}, vol.~34, no.~10, pp. 1992--2004, 2012.

\bibitem{hu2011sparse}
------, ``Sparse approximated nearest points for image set classification,'' in
  \emph{CVPR 2011}.\hskip 1em plus 0.5em minus 0.4em\relax IEEE, 2011, pp.
  121--128.

\bibitem{yamaguchi1998face}
O.~Yamaguchi, K.~Fukui, and K.-i. Maeda, ``Face recognition using temporal
  image sequence,'' in \emph{Proceedings Third IEEE International Conference on
  Automatic Face and Gesture Recognition}.\hskip 1em plus 0.5em minus
  0.4em\relax IEEE, 1998, pp. 318--323.

\bibitem{kim2007discriminative}
T.-K. Kim, J.~Kittler, and R.~Cipolla, ``Discriminative learning and
  recognition of image set classes using canonical correlations,'' \emph{IEEE
  Transactions on Pattern Analysis and Machine Intelligence}, vol.~29, no.~6,
  pp. 1005--1018, 2007.

\bibitem{wang2008manifold}
R.~Wang, S.~Shan, X.~Chen, and W.~Gao, ``Manifold-manifold distance with
  application to face recognition based on image set,'' in \emph{2008 IEEE
  Conference on Computer Vision and Pattern Recognition}.\hskip 1em plus 0.5em
  minus 0.4em\relax IEEE, 2008, pp. 1--8.

\bibitem{wang2009manifold}
R.~Wang and X.~Chen, ``Manifold discriminant analysis,'' in \emph{2009 IEEE
  Conference on Computer Vision and Pattern Recognition}.\hskip 1em plus 0.5em
  minus 0.4em\relax IEEE, 2009, pp. 429--436.

\bibitem{ETH80}
B.~{Leibe} and B.~{Schiele}, ``Analyzing appearance and contour based methods
  for object categorization,'' in \emph{2003 IEEE Computer Society Conference
  on Computer Vision and Pattern Recognition, 2003. Proceedings.}, vol.~2,
  2003, pp. II--409.

\bibitem{Cifar10krizhevsky2009learning}
A.~Krizhevsky, G.~Hinton \emph{et~al.}, ``Learning multiple layers of features
  from tiny images,'' 2009.

\bibitem{MNISTlecun1998gradient}
Y.~LeCun, L.~Bottou, Y.~Bengio, and P.~Haffner, ``Gradient-based learning
  applied to document recognition,'' \emph{Proceedings of the IEEE}, vol.~86,
  no.~11, pp. 2278--2324, 1998.

\bibitem{russakovsky2015imagenet}
O.~Russakovsky, J.~Deng, H.~Su, J.~Krause, S.~Satheesh, S.~Ma, Z.~Huang,
  A.~Karpathy, A.~Khosla, M.~Bernstein \emph{et~al.}, ``Imagenet large scale
  visual recognition challenge,'' \emph{International journal of computer
  vision}, vol. 115, no.~3, pp. 211--252, 2015.

\bibitem{abadi2016tensorflow}
M.~Abadi, P.~Barham, J.~Chen, Z.~Chen, A.~Davis, J.~Dean, M.~Devin,
  S.~Ghemawat, G.~Irving, M.~Isard \emph{et~al.}, ``Tensorflow: A system for
  large-scale machine learning,'' in \emph{12th $\{$USENIX$\}$ symposium on
  operating systems design and implementation ($\{$OSDI$\}$ 16)}, 2016, pp.
  265--283.

\bibitem{nicolae2018adversarial}
M.-I. Nicolae, M.~Sinn, M.~N. Tran, B.~Buesser, A.~Rawat, M.~Wistuba,
  V.~Zantedeschi, N.~Baracaldo, B.~Chen, H.~Ludwig \emph{et~al.}, ``Adversarial
  robustness toolbox v1. 0.0,'' \emph{arXiv preprint arXiv:1807.01069}, 2018.

\bibitem{meng2017magnet}
D.~Meng and H.~Chen, ``Magnet: a two-pronged defense against adversarial
  examples,'' in \emph{Proceedings of the 2017 ACM SIGSAC conference on
  computer and communications security}, 2017, pp. 135--147.

\bibitem{yang2019meNet}
Y.~Yang, G.~Zhang, D.~Katabi, and Z.~Xu, ``Me-net: Towards effective
  adversarial robustness with matrix estimation,'' \emph{arXiv preprint
  arXiv:1905.11971}, 2019.

\end{thebibliography}

%

\begin{IEEEbiography}[{\includegraphics[width=1in,height=1.25in,clip,keepaspectratio]{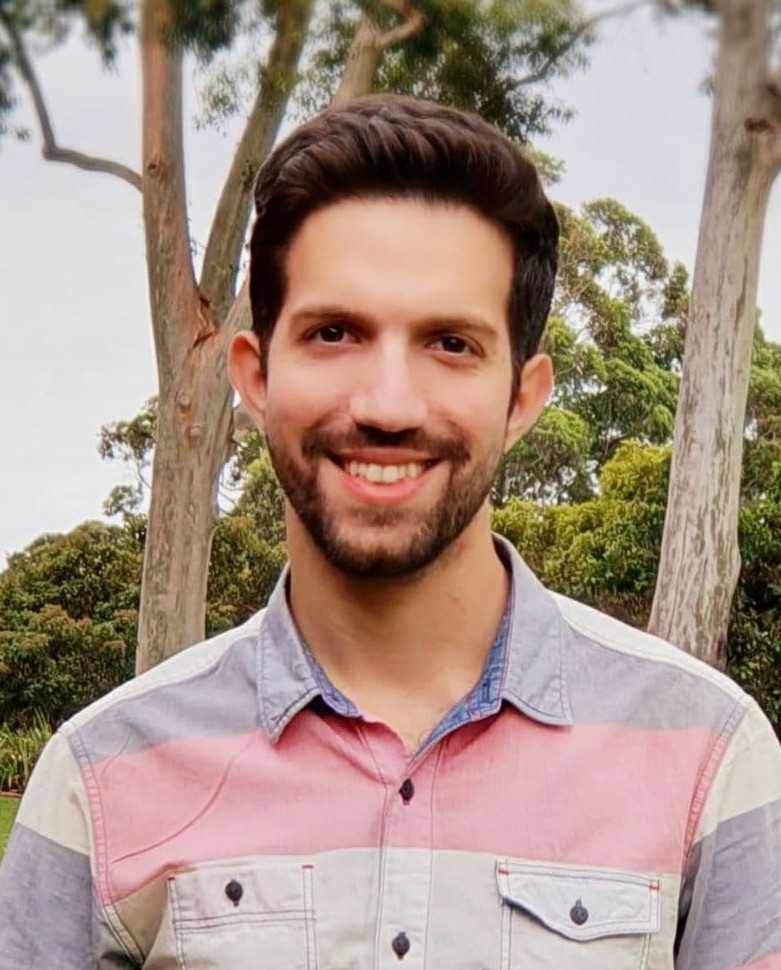}}]{Nima Mirnateghi}
is currently completing Master of Information Technology at Murdoch University, Perth, WA, Australia.
He received his Bachelor of Computer Science from the University of Wollongong (UOW). His current research interests include deep learning, pattern recognition, object recognition, facial detection, machine learning, and image processing. 
\end{IEEEbiography}


\begin{IEEEbiography}[{\includegraphics[width=1in,height=1.25in,clip,keepaspectratio]{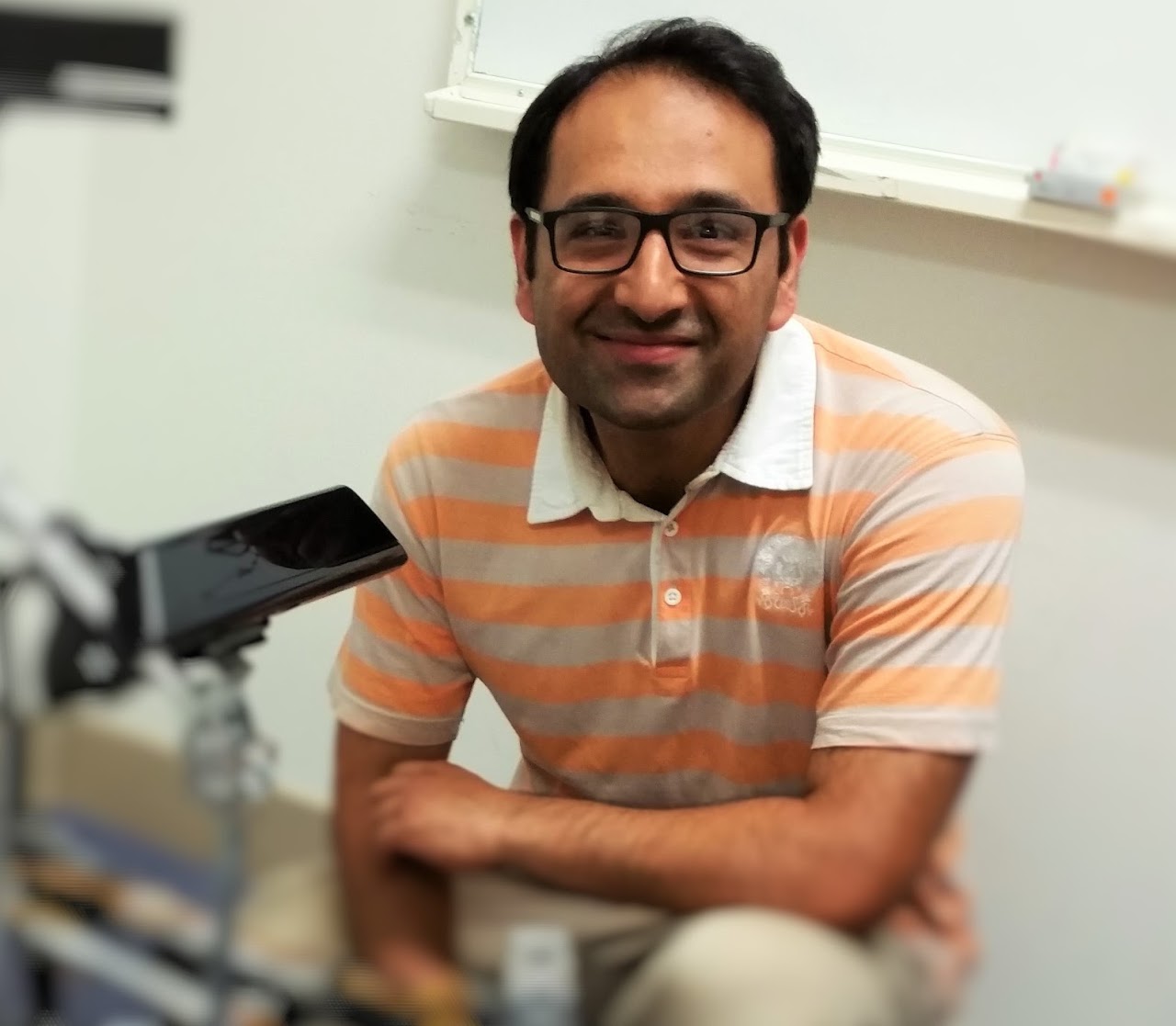}}]{Syed Afaq Ali Shah}
received the PhD degree in computer vision and machine learning from The University of Western Australia (UWA), Crawley, WA, Australia. He was a Lecturer ICT with Central Queensland University, Australia. He is currently a Lecturer at Murdoch University, Perth, WA, Australia. He is also an Adjunct Lecturer with the Department of Computer Science and Software Engineering, UWA, Perth, WA, Australia. His current research interests include deep learning, object/face recognition, Scene understanding, and image processing. Dr. Shah was a recipient of the Start Something Prize for Research Impact through Enterprise for 3-D Facial Analysis Project funded by the Australian Research Council. He has authored over 40 research articles and co-authored a book, A guide to convolutional neural networks for computer vision.
\end{IEEEbiography}

\begin{IEEEbiography}[{\includegraphics[width=1in,height=1.25in,clip,keepaspectratio]{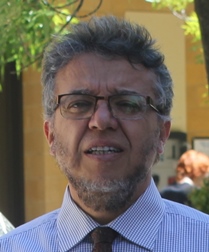}}]{Mohammed Bennamoun}
is Winthrop Professor in the Department of Computer Science and Software Engineering at UWA and is a researcher in computer vision, machine/deep learning, robotics, and signal/speech processing. He has published 4 books (available on Amazon, 1 edited book, 1 Encyclopedia article, 14 book chapters, 150+ journal papers, 250+ conference publications, 16 invited \& keynote publications. His h-index is 58 and his number of citations is 14,000+ (Google Scholar). He was awarded 65+ competitive research grants, from the Australian Research Council, and numerous other Government, UWA and industry Research Grants. He successfully supervised
28+ PhD students to completion. He won the Best Supervisor of the Year Award at QUT (1998), and received award for research supervision at UWA (2008 \& 2016) and Vice-Chancellor Award for mentorship (2016). He delivered conference tutorials at major conferences, including: IEEE Computer Vision and Pattern Recognition (CVPR 2016), Interspeech 2014, IEEE International Conference on Acoustics Speech and Signal Processing (ICASSP) and European Conference on Computer Vision (ECCV). He was also invited to give a Tutorial at an International Summer School on Deep Learning (DeepLearn 2017).
\end{IEEEbiography}




\setcounter{table}{1}
\onecolumn
\section*{Supplementary Material}
In supplementary material, we provide our complete experimental results for MNIST, Tiny ImageNet and ETH-80 datasets. The experiments are performed for 50 stochastic forward passes ($T = 50$), and different fractions of attack ratio $R$.
\begin{table}[h]
\renewcommand{\arraystretch}{1.12}
\caption{Performance Evaluation of The Proposed Technique on MNIST Data Set Against FGSM, PGD, and DeepFool Attacks.}
\label{MNIST_Table_All}
\centering
\begin{tabular}{|c|c|c|c|c|c|c|c|c|c|}
\hline
Model & Attack Type  & Adversarial Training & Perturbation Magnitude ($\epsilon$) & Attack Ratio (\%) & SV & MV & EWV \\
\hline
Baseline & FGSM  & No & -     & 0\%     & \textbf{100.00\%} & \textbf{100.00\%} & \textbf{100.00\%} \\
Baseline & FGSM  & No & 0.3   & 50\%   & 95.23\% & 96.19\% & 96.19\% \\
Baseline & FGSM  & No & 0.3   & 80\%   & 28.57\% & 28.57\% & 29.52\% \\
Baseline & FGSM  & No & 0.3   & 100\%   & 18.09\% & 18.09\% & 19.04\% \\
      &       &       &       &       &       &       &  \\
Baseline & FGSM  & Yes & -     & 0\%     & \textbf{100.00\%} & \textbf{100.00\%} & \textbf{100.00\%} \\
Baseline & FGSM  & Yes & 0.3   & 50\%   & \textbf{100.00\%} & \textbf{100.00\%} & \textbf{100.00\%} \\
Baseline & FGSM  & Yes & 0.3   & 80\%   & \textbf{100.00\%} & \textbf{100.00\%} & \textbf{100.00\%} \\
Baseline & FGSM  & Yes & 0.3   & 100\%   & \textbf{100.00\%} & \textbf{100.00\%} & \textbf{100.00\%} \\
      &       &       &       &       &       &       &  \\
Proposed & FGSM  & No & -     & 0\%     & \textbf{100.00\%} & \textbf{100.00\%} & \textbf{100.00\%} \\
Proposed & FGSM  & No & 0.3   & 50\%   & \textbf{100.00\%} & \textbf{100.00\%} & \textbf{100.00\%} \\
Proposed & FGSM  & No & 0.3   & 80\%   & 80.95\% & 77.14\% & 80.95\% \\
Proposed & FGSM  & No & 0.3   & 100\%     & 53.33\% & 54.28\% & 53.33\% \\
      &       &       &       &       &       &       &  \\
Proposed & FGSM  & Yes & -     & 0\%     & \textbf{100.00\%} & \textbf{100.00\%} & \textbf{100.00\%} \\
Proposed & FGSM  & Yes & 0.3   & 50\%   & \textbf{100.00\%} & \textbf{100.00\%} & \textbf{100.00\%} \\
Proposed & FGSM  & Yes & 0.3   & 80\%   & \textbf{100.00\%} & \textbf{100.00\%} & \textbf{100.00\%} \\
Proposed & FGSM  & Yes & 0.3   & 100\%   & \textbf{100.00\%} & \textbf{100.00\%} & \textbf{100.00\%} \\ \hline 

Baseline & PGD    & No & -     & 0\%     & \textbf{100.00\%} & \textbf{100.00\%} & \textbf{100.00\%} \\
Baseline & PGD    & No & 0.1   & 50\%   & \textbf{100.00\%} & \textbf{100.00\%} & \textbf{100.00\%} \\
Baseline & PGD    & No & 0.1   & 80\%   & 96.19\% & 95.23\% & 96.19\% \\
Baseline & PGD    & No & 0.1   & 100\%     & 90.47\% & 90.47\% & 90.47\% \\
      &              &       &       &       &       &       &  \\
Baseline & PGD    &    Yes & -     & 0\%     & \textbf{100.00\%} & \textbf{100.00\%} & \textbf{100.00\%} \\
Baseline & PGD    &   Yes & 0.1   & 50\%   & \textbf{100.00\%} & \textbf{100.00\%} & \textbf{100.00\%} \\
Baseline & PGD    &    Yes & 0.1   & 80\%   & \textbf{100.00\%} & \textbf{100.00\%} & \textbf{100.00\%} \\
Baseline & PGD    &   Yes & 0.1   & 100\%     & \textbf{100.00\%} & \textbf{100.00\%} & \textbf{100.00\%} \\
      &       &             &       &       &       &       &  \\
Proposed & PGD    & No & -     & 0\%     & \textbf{100.00\%} & \textbf{100.00\%} & \textbf{100.00\%} \\
Proposed & PGD    & No & 0.1   & 50\%   & \textbf{100.00\%} & \textbf{100.00\%} & \textbf{100.00\%} \\
Proposed & PGD   & No & 0.1   & 80\%   & \textbf{100.00\%} & \textbf{100.00\%} & \textbf{100.00\%} \\
Proposed & PGD   & No & 0.1   & 100\%     & \textbf{100.00\%} & \textbf{100.00\%} & \textbf{100.00\%} \\
      &              &       &       &       &       &       &  \\
Proposed & PGD    & Yes & -     & 0\%     & \textbf{100.00\%} & \textbf{100.00\%} & \textbf{100.00\%} \\
Proposed & PGD    & Yes & 0.1   & 50\%   & \textbf{100.00\%} & \textbf{100.00\%} & \textbf{100.00\%} \\
Proposed & PGD    & Yes & 0.1   & 80\%   & \textbf{100.00\%} & \textbf{100.00\%} & \textbf{100.00\%} \\
Proposed & PGD    & Yes & 0.1   & 100\%     & \textbf{100.00\%} & \textbf{100.00\%} & \textbf{100.00\%}  \\ 
\hline

Baseline & DeepFool & No & - & 0\%     & \textbf{100\%} & \textbf{100\%} & \textbf{1000\%} \\
Baseline & DeepFool & No & -  & 50\%    & \textbf{100\%} & \textbf{100\%} & \textbf{100\%} \\
Baseline & DeepFool & No & - & 80\%    & 23.80\%  & 24.76\%  & 25.71\% \\
Baseline & DeepFool & No & - & 100\%      & 17.14\%  & 18.10\%  & 18.10\% \\
      &       &       &    &   &       &       &  \\
Baseline & DeepFool & Yes & - & 0\%      & \textbf{100\%} & \textbf{100\%} & \textbf{100\%} \\
Baseline & DeepFool & Yes & - & 50\%    & \textbf{100\%} & \textbf{100\%} & \textbf{100\%} \\
Baseline & DeepFool & Yes & - & 80\%    & 87.61\%  & 86.66\%  & 87.61\% \\
Baseline & DeepFool & Yes & - & 100\%      & 64.76\%  & 60.95\%  & 64.76\% \\
      &       &       &   &    &       &       &  \\
Proposed & DeepFool & No & - & 0\%      & \textbf{100\%} & \textbf{100\%} & \textbf{100\%} \\
Proposed & DeepFool & No & - & 50\%    & \textbf{100\%} & \textbf{100\%} & \textbf{100\%} \\
Proposed & DeepFool & No & - & 80\%    & 31.42\%  & 30.47\%  & 32.38\% \\
Proposed & DeepFool & No & - & 100\%      & 22.85\%  & 23.80\%  & 22.85\% \\
      &       &       &  &     &       &       &  \\
Proposed & DeepFool & Yes & - & 0\%     & 94.28\%  & 91.42\%  & 94.28\% \\
Proposed & DeepFool & Yes & - & 50\%   & \textbf{100\%} & \textbf{100\%} & \textbf{100\% }\\
Proposed & DeepFool & Yes & - & 80\%   & 93.33\%  & 87.61\%  & 95.23\% \\
Proposed & DeepFool & Yes & - & 100\%     & 79.04\%  & 78.09\%  & 80.00\% \\

\hline
\end{tabular}%
\end{table}%

\begin{table*}[t]
\renewcommand{\arraystretch}{1.3}
\caption{Performance Evaluation of The Proposed Technique on Tiny ImageNet Data Set Against FGSM, PGD, and DeepFool Attacks}
\label{Tiny_ImageNet_Table_All}
\centering
\begin{tabular}{|c|c|c|c|c|c|c|c|c|c|}
\hline
Model & Attack Type  & Adversarial Training & Perturbation Magnitude ($\epsilon$) & Attack Ratio (\%) & SV & MV & EWV \\
\hline
Baseline & FGSM  & No & -     & 0\%     & 98.16\% & 96.66\% & 97.66\% \\
Baseline & FGSM  & No & 0.3   & 50\%   & 53\%  & 48.33\% & 55.66\% \\
Baseline & FGSM  & No & 0.3   & 80\%   & 3.33\% & 2.00\% & 3.66\% \\
Baseline & FGSM  & No & 0.3   & 100\%   & 1.16\% & 1.50\% & 1.00\% \\
      &       &       &       &       &       &       &  \\
Baseline & FGSM  & Yes & -     & 0\%    & 98.50\% & 96.66\% & 98.83\% \\
Baseline & FGSM  & Yes & 0.3   & 50\%   & 97.16\% & 92.00\% & 96.83\% \\
Baseline & FGSM  & Yes & 0.3   & 80\%  & 88.66\% & 79.66\% & 88.33\% \\
Baseline & FGSM  & Yes & 0.3   & 100\%  & 61.16\% & 50.50\% & 61.50\% \\
      &       &       &       &       &       &       &  \\
Proposed & FGSM  & No & -     & 0\%    & \textbf{99.66\%} & 99.50\% & \textbf{99.66\%} \\
Proposed & FGSM  & No & 0.3   & 50\%   & 17.33\% & 16.33\% & 20.16\% \\
Proposed & FGSM  & No & 0.3   & 80\%  & 0.83\% & 0.83\% & 0.83\% \\
Proposed & FGSM  & No & 0.3   & 100\%     & 0.66\% & 0.66\% & 0.66\% \\
      &       &       &       &       &       &       &  \\
Proposed & FGSM  & Yes & -     & 0\%    & 99.33\% & 98.33\% & 99.33\% \\
Proposed & FGSM  & Yes & 0.3   & 50\%   & 98.16\% & 96.50\% & 98.33\% \\
Proposed & FGSM  & Yes & 0.3   & 80\%   & 95.83\% & 92.16\% & 95.33\% \\
Proposed & FGSM  & Yes & 0.3   & 100\%  & 90.16\% & 84.50\% & 90.16\% \\
\hline
Baseline & PGD   &  No & -     & 0\%     & 98.17\% & 96.83\% & 97.66\% \\
Baseline & PGD   &  No & 0.03  & 50\%   & 94.33\% & 93.33\% & 93.33\% \\
Baseline & PGD   &  No & 0.03  & 80\%   & 84.00\% & 81.83\% & 81.16\% \\
Baseline & PGD   &  No & 0.03  & 100\%     & 64.66\% & 57.33\% & 61.83\% \\
      &       &             &       &       &       &       &  \\
Baseline & PGD   &  Yes & -     & 0\%     & 95.66\% & 92.33\% & 95.50\% \\
Baseline & PGD   &  Yes & 0.03  & 50\%   & 92.00\% & 86.83\% & 91.66\% \\
Baseline & PGD   &  Yes & 0.03  & 80\%   & 90.00\% & 84.16\% & 90.83\% \\
Baseline & PGD   &  Yes & 0.03  & 100\%     & 90.16\% & 85.16\% & 89.50\% \\
      &       &            &       &       &       &       &  \\
Proposed & PGD   & No & -     & 0\%     & \textbf{99.66\%} & 99.50\% & \textbf{99.66\%} \\
Proposed & PGD   & No & 0.03  & 50\%   & 99.33\% & 98.83\% & 99.16\% \\
Proposed & PGD   & No & 0.03  & 80\%   & 98.00\% & 97.00\% & 97.50\% \\
Proposed & PGD   & No & 0.03  & 100\%     & 97.00\% & 96.33\% & 96.33\% \\
      &       &           &       &       &       &       &  \\
Proposed & PGD   &  Yes & -     & 0\%     & 99.33\% & 98.66\% & 99.33\% \\
Proposed & PGD   & Yes & 0.03  & 50\%   & 98.50\% & 97.66\% & 98.33\% \\
Proposed & PGD   &  Yes & 0.03  & 80\%   & 98.50\% & 98.50\% & 98.33\% \\
Proposed & PGD   &  Yes & 0.03  & 100\%     & 99.00\% & 98.00\% & 98.50\% \\
\hline
Baseline & DeepFool & No    & -     & 0\%   & 98.17\% & 96.83\% & 97.66\% \\
Baseline & DeepFool & No    & -     & 50\%  & 94.83\% & 91.33\% & 95.33\% \\
Baseline & DeepFool & No    & -     & 80\%  & 92.50\% & 89.33\% & 93.50\% \\
Baseline & DeepFool & No    & -     & 100\% & 86.66\% & 79.50\% & 86.16\% \\
      &       &       &       &       &       &       &  \\
Baseline & DeepFool & Yes   & -     & 0\%   & 96.83\% & 95.33\% & 87.17\% \\
Baseline & DeepFool & Yes   & -     & 50\%  & 92.83\% & 88.16\% & 92.66\% \\
Baseline & DeepFool & Yes   & -     & 80\%  & 93.50\% & 89.00\% & 92.33\% \\
Baseline & DeepFool & Yes   & -     & 100\% & 91.33\% & 87.17\% & 91.33\% \\
      &       &       & -     &       &       &       &  \\
Proposed & DeepFool & No    & -     & 0\%   & \textbf{99.66\%} & 99.50\% & \textbf{99.66\%} \\
Proposed & DeepFool & No    & -     & 50\%  & 99.50\% & 99.33\% & 99.50\% \\
Proposed & DeepFool & No    & -     & 80\%  & 99.33\% & 98.00\% & 99.33\% \\
Proposed & DeepFool & No    & -     & 100\% & 97.66\% & 96.00\% & 98.16\% \\
      &       &       &       &       &       &       &  \\
Proposed & DeepFool & Yes   & -     & 0\%   & 99.33\% & 99.16\% & 99.33\% \\
Proposed & DeepFool & Yes   & -     & 50\%  & 99.00\% & 98.33\% & 99.00\% \\
Proposed & DeepFool & Yes   & -     & 80\%  & 99.16\% & 98.33\% & 99.17\% \\
Proposed & DeepFool & Yes   & -     & 100\% & 99.50\% & 98.66\% & 99.16\% \\

\hline
\end{tabular}%
\end{table*}%
\begin{table*}[t]
\renewcommand{\arraystretch}{1.3}
\caption{Performance Evaluation of The Proposed Technique on ETH-80 Data Set Against FGSM, PGD, and DeepFool Attacks}
\label{ETH_Table_All}
\centering
\begin{tabular}{|c|c|c|c|c|c|c|c|c|c|}
\hline
Model & Attack Type & Split Ratio  & Adversarial Training & Perturbation Magnitude ($\epsilon$) & Attack Ratio (\%) & SV & MV & EWV \\
\hline
Baseline & FGSM  & 80-20 & No & -     & 0\%     & 90.00\% & 87.50\% & 87.50\% \\
Baseline & FGSM  & 80-20 & No & 0.3   & 50\% & 78.75\% & 80.00\% & 80.00\% \\
Baseline & FGSM  & 80-20 & No & 0.3   & 80\% & 27.50\% & 40.00\% & 41.25\% \\
Baseline & FGSM  & 80-20 & No & 0.3   & 100\% & 25.00\% & 25.00\% & 25.00\% \\
      &       &       &       &       &       &       &       &  \\
Baseline & FGSM  & 80-20 & Yes & -     & 0\%  & 98.75\% & 98.75\% & 98.75\% \\
Baseline & FGSM  & 80-20 & Yes & 0.3   & 50\% & 96.25\% &  97.50\% & 98.75\% \\
Baseline & FGSM  & 80-20 & Yes & 0.3   & 50\% & 87.50\% & 96.25\% & 96.25\% \\
Baseline & FGSM  & 80-20 & Yes & 0.3   & 100\% & 81.25\% & 83.75\% & 85.00\% \\
    &       &       &       &       &       &       &       &  \\
Proposed & FGSM  & 80-20 & No & -     & 0\%  & 97.50\% & 98.75\% & 98.75\% \\
Proposed & FGSM  & 80-20 & No & 0.3   & 50\% &82.25\% & 85.00\% & 83.75\% \\
Proposed & FGSM  & 80-20 & No & 0.3   & 50\%& 12.50\% & 12.50\% & 12.50\% \\
Proposed & FGSM  & 80-20 & No & 0.3   & 100\% & 12.50\% & 12.50\% & 12.50\% \\
      &       &       &       &       &       &       &       &  \\
Proposed & FGSM  & 80-20 & Yes & -     & 0\%  & 91.25\% & 86.25\% & 86.25\% \\
Proposed & FGSM  & 80-20 & Yes & 0.3   & 50\% & 87.50\% & 87.50\% & 87.50\% \\
Proposed & FGSM  & 80-20 & Yes & 0.3   & 50\% & 85.00\% & 85.00\% & 83.75\% \\
Proposed & FGSM  & 80-20 & Yes & 0.3   & 100\%& 78.75\% & 78.75\% & 80.00\% \\
\hline
Baseline & PGD   & 80-20 & No & -     & 0\%      & 90.00\% & 87.50\% & 87.50\% \\
Baseline & PGD   & 80-20 & No & 0.3   & 50\%   & 80.00\% & 75.00\% & 76.25\% \\
Baseline & PGD   & 80-20 & No & 0.3   & 50\%   & 6.25\% & 6.25\% & 6.25\% \\
Baseline & PGD   & 80-20 & No & 0.3   & 100\%     & 3.75\% & 3.75\% & 3.75\% \\
      &       &       &       &       &       &       &       &  \\
Baseline & PGD   & 80-20 & Yes & -     & 0\%      & 95.00\% & 95.00\% & 95.00\% \\
Baseline & PGD   & 80-20 & Yes & 0.3   & 50\%   & 96.25\% & 96.25\% & 96.25\% \\
Baseline & PGD   & 80-20 & Yes & 0.3   & 50\%   & 92.50\% & 98.75\% & 97.50\% \\
Baseline & PGD   & 80-20 & Yes & 0.3   & 100\%     & 88.75\% & 85.00\% & 86.25\% \\
      &       &       &       &       &       &       &       &  \\
Proposed & PGD   & 80-20 & No & -     & 0\%     & 97.50\% & 98.75\% & 98.75\% \\
Proposed & PGD   & 80-20 & No & 0.3   & 50\%   & 82.50\% & 85.00\% & 83.75\% \\
Proposed & PGD   & 80-20 & No & 0.3   & 50\%   & 12.50\% & 12.50\% & 12.50\% \\
Proposed & PGD   & 80-20 & No & 0.3   & 100\%     & 12.50\% & 12.50\% & 12.50\% \\
      &       &       &       &       &       &       &       &  \\
Proposed & PGD   & 80-20 & Yes & -     & 0\%      & 95.00\% & 95.00\% & 96.25\% \\
Proposed & PGD   & 80-20 & Yes & 0.3   & 50\%   & 93.75\% & 98.75\% & 98.75\% \\
Proposed & PGD   & 80-20 & Yes & 0.3   & 50\%   & 77.50\% & 80.00\% & 78.75\% \\
Proposed & PGD   & 80-20 & Yes & 0.3   & 100\%     & 72.50\% & 72.50\% & 73.75\% \\
\hline
Baseline & DeepFool & 80-20 & No & - & 0\%      & 90.00\% & 87.50\% & 87.50\% \\
Baseline & DeepFool & 80-20 & No & - & 50\%   & 86.25\% & 82.50\% & 82.50\% \\
Baseline & DeepFool & 80-20 & No & - & 50\%   & 31.25\% & 43.75\% & 50.00\% \\
Baseline & DeepFool & 80-20 & No & - & 100\%     & 26.25\% & 22.50\% & 23.75\% \\
      &       &       &       &       &   &    &       &  \\
Baseline & DeepFool & 80-20 & Yes & - & 0\%      & \textbf{100.00\%} & \textbf{100.00\%} & \textbf{100.00\%} \\
Baseline & DeepFool & 80-20 & Yes & - & 50\%   & 93.75\% & 93.75\% & 95.00\% \\
Baseline & DeepFool & 80-20 & Yes & - & 50\%   & 91.25\% & 92.50\% & 92.50\% \\
Baseline & DeepFool & 80-20 & Yes & - & 100\%     & 88.75\% & 92.50\% & 91.25\% \\
      &       &       &       &       &    &   &       &  \\
Proposed & DeepFool & 80-20 & No & - & 0\%      & 97.50\% & 98.75\% & 98.75\% \\
Proposed & DeepFool & 80-20 & No & - & 50\%   & 93.75\% & 95.00\% & 95.00\% \\
Proposed & DeepFool & 80-20 & No & - & 50\%   & 62.50\% & 63.75\% & 71.25\% \\
Proposed & DeepFool & 80-20 & No & - & 100\%     & 56.25\% & 55.00\% & 60.00\% \\
      &       &       &       &       &    &   &       &  \\
Proposed & DeepFool & 80-20 & Yes & - & 0\%      & \textbf{100.00\%} & \textbf{100.00\%} &\textbf{100.00\%}\\
Proposed & DeepFool & 80-20 & Yes & - & 50\%   & \textbf{100.00\%} & \textbf{100.00\%} & \textbf{100.00\%} \\
Proposed & DeepFool & 80-20 & Yes & - & 50\%   & 96.25\% & 96.25\% & 95.00\% \\
Proposed & DeepFool & 80-20 & Yes & - & 100\%     & 93.75\% & 95.00\% & 95.00\% \\
\hline

\end{tabular}%
\end{table*}%

\end{document}